%% file: main.tex
\documentclass{article}
\usepackage{geometry}

\usepackage[utf8]{inputenc} % allow utf-8 input
\usepackage[T1]{fontenc}    % use 8-bit T1 fonts
\usepackage[pdflang={en-US}]{hyperref}       % hyperlinks
\usepackage{url}            % simple URL typesetting
\usepackage{booktabs}       % professional-quality tables
\usepackage{amsfonts}       % blackboard math symbols
\usepackage{nicefrac}       % compact symbols for 1/2, etc.
\usepackage{microtype}      % microtypography
\usepackage{xcolor}         % colors

\usepackage[utf8]{inputenc}
\usepackage{graphicx}
\usepackage{mathtools}
\usepackage{pdflscape}
\usepackage{listings}
\usepackage{caption}
\usepackage{bbm}
\usepackage{amsthm}
\usepackage{amsmath}
\usepackage{amssymb}
\usepackage[square,sort,comma,numbers]{natbib}
\usepackage{subfig}
\usepackage{authblk}

\usepackage{algorithm}
\usepackage{algorithmic}

\newtheorem{lemma}{Lemma}
\newtheorem{theorem}{Theorem}
\newtheorem*{theorem*}{Theorem}
\newtheorem{remark}{Remark}

\newcommand{\E}{\mathbb{E}}
\newcommand{\cX}{\mathcal{X}}

\newcommand{\cH}{\mathcal{H}}
\newcommand{\R}{\mathbb{R}}

\newcommand{\bfx}{\mathbf{x}}
\newcommand{\bfy}{\mathbf{y}}

\newcommand{\N}{\mathbb{N}}
\newcommand{\ip}[1] {\left\langle #1 \right\rangle }
\newcommand{\term}{\mathrm{term}}
\newcommand{\cT}{\mathcal{T}}
\newcommand{\cI}{\mathcal{I}}
\newcommand{\cV}{\mathcal{V}}

\newcommand{\cL}{\mathcal{L}}
\newcommand{\cD}{\mathcal{D}}

\newcommand{\xd}{x^*_{D}}
\newcommand{\xdg}{x^*_{D_g}}
\newcommand{\cC}{\mathcal{C}}

\DeclareMathOperator*{\argmax}{arg\,max}

\usepackage{hyperref}  
\hypersetup{
    colorlinks=true,
    linkcolor=blue,
    citecolor =blue,
    filecolor=magenta,      
    urlcolor=magenta,
}

\title{A Domain-Shrinking based Bayesian Optimization Algorithm with Order-Optimal Regret Performance}
\author[$\ddagger$]{Sudeep Salgia}
\author[*]{Sattar Vakili}
\author[$\ddagger$]{Qing Zhao}
\affil[$\ddagger$]{School of Electrical \& Computer Engineering, Cornell University, Ithaca, NY, \emph{\{ss3827,qz16\}@cornell.edu} }
\affil[*]{MediaTek Research, UK, \emph{sattar.vakili@mtkresearch.com}}

\date{}

\begin{document}

\maketitle

\begin{abstract}
We consider sequential optimization of an unknown function in a reproducing kernel Hilbert space. We propose a Gaussian process-based algorithm and establish its order-optimal regret performance (up to a poly-logarithmic factor). This is the first GP-based algorithm with an order-optimal regret guarantee. The proposed algorithm is rooted in the methodology of domain shrinking realized through a sequence of tree-based region pruning and refining to concentrate queries in increasingly smaller high-performing regions of the function domain. The search for high-performing regions is localized and guided by an iterative estimation of the optimal function value to ensure both learning efficiency and computational efficiency. Compared with the prevailing GP-UCB family of algorithms, the proposed algorithm reduces computational complexity by a factor of $O(T^{2d-1})$ (where $T$ is the time horizon and $d$ the dimension of the function domain).
\end{abstract}

\input{introduction}

\input{problem_formulation}
\input{algorithm_description}
\input{analysis}

\input{simulations}
\input{conclusion}

\bibliography{citations}
\bibliographystyle{unsrt}

\input{appendix}

%%%%%%%%%%%%%%%%%%%%%%%%%%%%%%%%%%%%%%%%%%%%

\end{document}

%% file: introduction.tex
\section{Introduction} % (fold)
\label{sec:introduction}

Consider a black-box optimization problem with an unknown objective function $f: \cX \to \R$, where $\cX \subset \R^d$ is a convex and compact set.  The learner can access the function only through a noisy oracle, which, when queried with a point $x \in \cX$, returns a noisy function value at that point. The learning objective is to approach the maximizer $x^*$ of the function through a sequence of query points $\{x_t\}_{t = 1}^T$ chosen sequentially in time. The learning efficiency is measured by cumulative regret given by
\begin{align}
	R(T) = \sum_{t = 1}^T \left[ f(x^*) - f(x_t) \right]. 
	\label{eqn:regret_definition}
\end{align}
This cumulative regret measure dictates the online nature of the problem: every query point during the learning process carries loss, not just the end point $x_T$ after learning concludes. The classical exploration-exploitation tradeoff in online learning hence ensues.

\subsection{Gaussian Process Models} % fold
\label{sub:GP_models}

The above problem is ill-posed unless certain structure of the unknown objective function $f$ is assumed to make learning $x^*$ feasible. One such structural assumption is the convexity of $f$, which leads to the class of stochastic convex optimization problems. Another class of black-box optimization problems that is gaining interest in recent years is kernel-based learning where $f$ is assumed to live in a Reproducing Kernel Hilbert Space (RKHS) associated with a positive-definite kernel. An effective approach to kernel-based black-box optimization is Bayesian optimization that adopts a \emph{fictitious} prior on the unknown function $f$. In other words, while $f$ is deterministic, it is viewed internally by the learning algorithm as a realization of a random process over $\cX$. A natural choice is the Gaussian process (GP) with a Gaussian prior due to the conjugate property that significantly simplifies the analytical form of the posterior distribution at each newly obtained observation.   \\

In a celebrated work, Srinivas \emph{et al.}~\cite{Srinivas2010} proposed the GP-UCB algorithm that constructs a proxy of $f$ using the upper confidence bound (UCB) concept first introduced in the classical multi-armed bandit problem~\cite{Lai1985, Auer2002}. Specifically, at each time instant $t$, a UCB of $f$ is constructed using the closed-form posterior mean and standard deviation of the GP model of $f$. The algorithm then sets the next query point to be the maximizer of the UCB.  Several variations of GP-UCB, tailored for different settings (see Sec~\ref{sub:related_work}), have since been developed. \\ 
% e.g., parallelization and multi-fidelity evaluations; 

The GP-UCB family of algorithms generally enjoy good empirical performance in terms of regret. The analytical guarantees of their regret performance, however, leave considerable gaps to
% There is, however, a significant gap between their regret upper bounds and 
the existing lower bound~\cite{Scarlett2017}. More significantly, the state-of-the-art regret bound of GP-UCB does not guarantee a sublinear order in $T$ for certain kernels, hence a lack of guaranteed convergence to $f(x^*)$~\cite{Scarlett2017, Janz2020}.   \\

Another difficulty with the GP-UCB family of algorithms is their computational complexity, which can be prohibitive as the dimension $d$ and/or the horizon length $T$ grows.  %high-dimensional problems or a  long learning horizon. 
The computational complexity has two main sources: (i) the inversion of the covariance matrix in updating the posterior GP distribution, which has an $O(t^3)$ complexity with $t$ samples; (ii) the maximization of the UCB proxy over the entire domain $\cX$ at each time instant. In particular, due to the multi-modality of the UCB score, its maximization is often carried out using a grid search with an increasingly finer discretization of the entire domain. %to ensure convergence to the optimal point.
Specifically, due to analytical requirements, the discretization is typically assumed to grow in the order of $O(t^{2d})$~\cite{Srinivas2010, Chowdhury2017}, resulting in an overall computational complexity of $O(T^{2d + 3})$. \\
% the discretization  grows in the order of $O(t^{2d})$~\cite{Srinivas2010, Chowdhury2017}, resulting in an overall computational complexity of $O(T^{2d + 3})$. 

Several studies exist that tackle the first source of high complexity of GP-UCB, using sparse matrix approximation techniques to reduce the complexity in the inversion of the covariance matrix (see, e.g., \cite{McWilliams2013, Calandriello2019}). The second source, which is the dominating factor, 
% source of computational complexity,
has not been effectively addressed. % and is the focus of this work. 

% subsection gaussian_process_models (end)

\subsection{Main results} % (fold)
\label{sub:main_results}

% The contribution of this work is a computationally efficient algorithm with a regret guarantee that closes the gap to the existing lower bound. 

The goal of this work is to develop a GP-based Bayesian optimization algorithm with a regret guarantee that closes the gap to the lower bound. Furthermore, we tackle the second source of the complexity to ensure both learning efficiency and computational efficiency.   \\

Referred to as GP-ThreDS (Thresholded Domain Shrinking), the proposed algorithm is rooted in the methodology of domain shrinking: it continuously prunes sub-performing regions of the domain $\cX$ and zooms into increasingly smaller high-performing regions of $\cX$ as time goes. The purpose of the domain shrinking is twofold. First, it ensures high learning efficiency by focusing queries on regions of $\mathcal{X}$ with function values approaching $f(x^*)$. Second, it achieves computational efficiency by avoiding a global maximization of the proxy function over the entire domain $\cX$.    \\

Our specific approach to domain shrinking is built upon a sequence of localized searches on a growing binary tree that forms successively refined partitions of $\cX$. Starting from the root of the tree that represents the entire domain, the search progresses down the tree by 
adaptively pruning nodes that do not contain the maximizer with high probability, consequently zooming into increasingly smaller high-performing regions of $\cX$ as the search deepens. Another progressive thread in this sequence of localized searches is the criterion for pruning the tree. Each localized search aims to identify nodes at a certain depth of the tree that contain points with function values exceeding a given threshold. The threshold is updated iteratively to approach the maximum function value $f(x^*)$. More succinctly, the proposed algorithm is a sequence of localized searches in the domain of the function guided by an iterative search in the range of the function. \\

The above domain shrinking approach via localized search is the primary contributing factor to improved performance in terms of both regret guarantee and computational complexity. In particular, the rate of domain shrinking is controlled to ensure not only the concentration of query points in high-performing regions, but also a \emph{constant}-sized discretization at all times when estimating the function values.
% guided by an iterative search in the range of the function, domain shrinking prunes the tree in a manner that allows the learner to focus queries on regions with high function values thereby ensuring learning efficiency. Furthermore, the rate of domain shrinking is tuned to ensure that the evaluation of the function proxies are carried out on a discretization of \emph{constant} size at all times. 
This constant-sized discretization allows a tighter regret analysis and results in a regret upper bound for GP-ThreDS that matches with the lower bound (up to a poly-logarithmic factor). We show that the regret of GP-ThreDS is $O(\sqrt{T \gamma_T})$ (up to a poly-logarithmic factor), where $\gamma_T$ denotes the maximum information gain after $T$ steps and is representative of the \emph{effective} dimension of the problem~\cite{Zhang2005, Valko2013a}.
In the case of Mat\'ern and Squared Exponential (SE) kernels where the lower bounds on regret are known, on substituting the improved bounds on $\gamma_T$ from~\cite{Vakili2020b}, our results match the lower bounds and close the gap reported in~\cite{Scarlett2017, Cai2020}. In comparison, the state-of-the-art analysis of GP-UCB yields an $O(\gamma_T\sqrt{T})$ regret bound~\citep[e.g., see,][Theorem $3$]{Chowdhury2017}. The $O(\sqrt{\gamma_T})$ gap between the regret guarantees of  GP-UCB and the proposed GP-ThreDS is significant: it can grow polynomially in $T$ (e.g. in the case of Mat\'ern kernel). \\

Computation-wise, the constant-sized discretization contrasts sharply with the growing (at rate $O(t^{2d})$ with time $t$) discretization required by the GP-UCB family of algorithms. 
% As a result, the proposed domain shrinking strategy significantly reduces the computational complexity while offering improved regret guarantees. 
Another factor contributing to the reduced complexity is the relaxed search criterion that aims to determine only the existence of threshold-exceeding points, in contrast to finding a global maximizer as in the GP-UCB family of algorithms. As a result, GP-ThreDS reduces the computational complexity from $O(T^{2d+3})$ as required by GP-UCB family of algorithms to $O(T^4)$.

\subsection{Related Work} % (fold)
\label{sub:related_work}

There is a vast body of literature on numerical and theoretical analysis of Bayesian optimization algorithms. With our focus on a computationally efficient algorithm with a provable regret guarantee, the most relevant results to ours are~\cite{Srinivas2010} and~\cite{Chowdhury2017} discussed above.
\cite{Chowdhury2017} also proved the same $O(\gamma_T\sqrt{T})$ regret holds for GP-TS, a Bayesian optimization algorithm based on Thompson sampling principle. 
Augmenting GP models with local polynomial estimators, \cite{Shekhar2020} introduced LP-GP-UCB and established improved regret bounds for it under special cases \cite[see,][Sec. $3.2$]{Shekhar2020}. However, for other cases, the regret guarantees for LP-GP-UCB remain in the same order as GP-UCB.
More recently,~\cite{Janz2020}
introduced $\pi$-GP-UCB, specific to Mat{\'e}rn family of kernels, that constructs a cover for the search space, as many hypercubes, and fits an independent GP to each cover element. This algorithm
was proven to achieve sublinear regret across all parameters of the Mat{\'e}rn family.
Almost all other algorithms in the GP-UCB family have a regret guarantee of $O(\gamma_T \sqrt{T})$, which is $O(\sqrt{\gamma_T})$ greater than the lower bound and can grow polynomially in $T$. Two exceptions to this are the SupKernelUCB and the RIPS algorithms proposed in~\cite{Valko2013a} and~\cite{Camilleri2021} which achieve a regret of $O(\sqrt{T \gamma_T})$ for discrete action spaces.%
% An exception is the SupKernelUCB algorithm proposed in~\cite{Valko2013a} which achieves a regret of $O(\sqrt{T \gamma_T})$ for discrete action spaces.
While this may be extendable to continuous spaces via a discretization argument as recently pointed out in~\cite{Janz2020, Cai2020}, the required discretization needs to grow polynomially in $T$, making it computationally expensive. Moreover, it has been noted that SupKernelUCB performs poorly in practice~\cite{ Janz2020,Calandriello2019, Cai2020}.
GP-ThreDS, on the other hand, is a computationally efficient algorithm that achieves tight regret bounds with good empirical performance (see Sec.~\ref{sec:simulations}).
A comparison with other related works including the ones in different settings such as noise-free observations and random $f$ are deferred to Appendix~\ref{sec:related_work}.

%% file: problem_formulation.tex
\section{Problem Statement} % (fold)
\label{sec:problem_setup}

\subsection{Problem Formulation}
\label{sub:problem_formualation}

We consider the problem of optimizing a fixed and unknown function $f: \cX \to \R$, where $\cX \subset \R^d$ is a convex and compact domain. A sequential optimization algorithm chooses a point $x_t \in \cX$ at each time instant $t = 1, 2, \dots, $ and observes $y_t = f(x_t) + \epsilon_t$, where the noise sequence $\{\epsilon_t\}_{t = 1}^{\infty}$ is assumed to be i.i.d. over $t$ and $R$-sub-Gaussian for a fixed constant $R \geq 0$, i.e., $\E \left[ e^{\zeta \epsilon_t} \right] \leq \exp \left( \zeta^2 R^2/2 \right)$
for all $\zeta \in \R$ and $t \in \N$.  \\

We assume a regularity condition on the objective function $f$ that is commonly adopted under kernelized learning models.
Specifically, we assume that $f$ lives in a Reproducing Kernel Hilbert Space (RKHS)%\footnote{Please refer to the supplementary material for a brief introduction to RKHS.}
\footnote{The RKHS, denoted by $H_k$, is a Hilbert space associated with a positive definite kernel $k(\cdot, \cdot)$ and is fully specified by the kernel and vice versa. It is endowed with an inner product $\ip{\cdot}_k$ that obeys the reproducing property, i.e., $g(x) = \ip{g, k(x, \cdot)}_k$ for all $g \in H_k$. The inner product also induces a norm $\|g \|_{k} = \ip{g, g}_k$. This norm is a measure of the smoothness of the function $f$ with respect to the kernel $k$ and is finite if and only if $f \in H_k$.} 
associated with a positive definite kernel $k: \cX \times \cX \to \R$. The RKHS norm of $f$ is assumed to be bounded by a known constant $B$, that is, $\|f \|_{k} \leq B$. 
We further assume that $f$ is $\alpha$-H{\"o}lder continuous, that is, $|f(x) - f(x')| \leq L \|x - x'\|^{\alpha}$ for all $x,x' \in \cX$ for some $\alpha \in (0, 1]$ and $L > 0$. This is a mild assumption as this is a direct consequence of RKHS assumption for commonly used kernels as shown in \cite{Shekhar2020}. We also assume the knowledge of an interval $[a,b]$, such that $f(x^*) \in [a,b]$. This is also a mild assumption as domain-specific knowledge often provides us with bounds. For example, a common application of black-box optimization is hyperparameter tuning in deep learning models. The unknown function represents the accuracy of the model for a given set of hyperparameters. Since $f$ represents the accuracy of the model, we have $f(x^*) \in [0,1]$. For simplicity of notation, we assume $\cX = [0,1]^d$ and $f(x^*) \in [0,1]$. It is straightforward to relax these assumptions to general compact domains and arbitrary bounded ranges $[a,b]$.  \\

Our objective is a computationally efficient algorithm with a guarantee on regret performance as defined in~\eqref{eqn:regret_definition}. We provide high probability regret bounds that hold with probability at least $1 - \delta_0$ for any given $\delta_0 \in (0,1)$, a stronger performance guarantee than bounds on expected regret.

\subsection{Preliminaries on Gaussian processes}
\label{sub:preliminaries_on_gaussian_processes}

Under the GP model, the unknown function $f$ is treated hypothetically as a realization of a Gaussian process over $\cX$.
A Gaussian Process $\{F(x)\}_{x \in \cX}$ is fully specified by its mean function $\mu(\cdot)$ and covariance function $k(\cdot, \cdot)$. All finite samples of the process are jointly Gaussian with mean $\E[F(x_i)] = \mu(x_i)$ and covariance $\E[(F(x_i) - \mu(x_i))(F(x_j) - \mu(x_j))] = k(x_i , x_j)$ for $1 \leq i, j\leq n$ and $n \in \N$~\cite{Rasmussen2005}. The noise $\epsilon_t$ is also viewed as Gaussian. \\

% Gaussian processes are closely related to RKHSs~\cite{Kanagawa2018}. In particular GPs provide powerful non-parametric Bayesian models over the space of functions which can be applied to RKHS elements.

The conjugate property of Gaussian processes with Gaussian noise allows for a closed-form expression of the posterior distribution. Consider a set of observations $\cH_t = \{\bfx_t, \bfy_t\}$ where $\bfx_t = (x_1, x_2, \dots, x_t)^T$ and $\bfy_t = (y_1, y_2, \dots, y_t)^T$. Here $y_s = {f}(x_s) + \epsilon_s$ where $x_s \in \cX$ and $\epsilon_s$ are the zero-mean noise terms,
i.i.d. over $s$ for $s \in \N$. Conditioned on the history of observations $\cH_t$, the posterior for $f$ is also a Gaussian process with mean and covariance functions given as %follows
\begin{align}
	\mu_t(x) & = \E \left[F(x) |\cH_t\right] = k_{\bfx_t, x}^T \left( K_{\bfx_t, \bfx_t} + \lambda I \right)^{-1} \bfy_t \label{eq:posterior_mean} \\
	k_t(x, x') & = \E\left[(F(x) - \mu_t(x))(F(x') - \mu_t(x'))|\cH_t\right]   = k(x, x') - k_{\bfx_t, x}^T \left( K_{\bfx_t, \bfx_t} + \lambda I \right)^{-1} k_{\bfx_t, x'}. \label{eq:posterior_variance}
\end{align}
In the above expressions, $k_{\bfx_t, x} = [ k(x_1, x), \dots, k(x_t, x) ]^T$, $K_{\bfx_t, \bfx_t}$ is the $t \times t$ covariance matrix  $[k(x_i, x_j)]_{i,j = 1}^t$, $I$ is the $t \times t$ identity matrix and $\lambda$ is the variance of the Gaussian model assumed for the noise terms.  \\
% The bottleneck associated with computing the posterior distribution lies in inverting the matrix $K_{\bfx_t, \bfX_t} + \lambda I $, a step that has an $O(t^3)$ computational complexity.

Gaussian processes are powerful non-parametric Bayesian models for functions in RKHSs~\cite{Kanagawa2018}. In particular, the mean function of the GP regression (eqn.~\eqref{eq:posterior_mean}) lies in the RKHS with kernel $k(\cdot, \cdot)$ with high probability. %Thereby, GPs provide powerful non-parametric Bayesian models for functions in RKHS.
We emphasize that the GP model of  $f$ and the Gaussian noise assumption are internal to the learning algorithm. The underlying objective function $f$ is an arbitrary deterministic function in an RKHS, and the noise obeys an arbitrary $R$-sub-Gaussian distribution.

% section problem_formulation (end)

%% file: algorithm_description.tex
\section{The GP-ThreDS Algorithm} % (fold)
\label{sec:algorithm_description}

In Sec.~\ref{sub:domain_shrinking}, we present the basic domain-shrinking structure of GP-ThreDS that continuously prunes sub-performing regions of $\cX$ and zooms into increasingly smaller high-performing regions of $\cX$. In Sec.~\ref{sub:random_walk_on_a_tree}, we present the method for identifying high-performing regions of $\cX$.

% We describe GP-ThreDS in two steps. In Sec.~\ref{sub:domain_shrinking}, we present the overall thresholded domain shrinking structure that traverses a binary tree representation of the function domain $\cX$ and progresses in epochs---super time units when regions of $\cX$ (i.e., nodes of the tree) are either pruned or refined/grown into their sub-regions. In Sec.~\ref{sub:random_walk_on_a_tree}, we present the detailed method for pruning/growing nodes on the tree within each epoch.    

% \begin{figure*}[t]
% \centering
% \subfloat[Thresholded domain shrinking: an example.]{\label{fig:thresholds}\centering \includegraphics[scale = 0.2]{thresholds.pdf}}
% ~~~~~
% \subfloat[A sample path of RWT algorithm. The random walk is initialized at the node indexed $1$ while $6$ denotes the target node. The local test module is first carried out on nodes $2$ and $3$ and based on the results the random walk moves to node $3$ (more likely outcome). Using a similar process, the random walk zooms into node $6$, where it performs the leaf level test and after obtaining a $+1$, the iteration terminates.]{\label{fig:RWT}\centering \includegraphics[scale = 0.35]{RWT.pdf}}
% \end{figure*}

\begin{figure}
\begin{minipage}{0.45\textwidth}
    \centering
    \includegraphics[scale = 0.2]{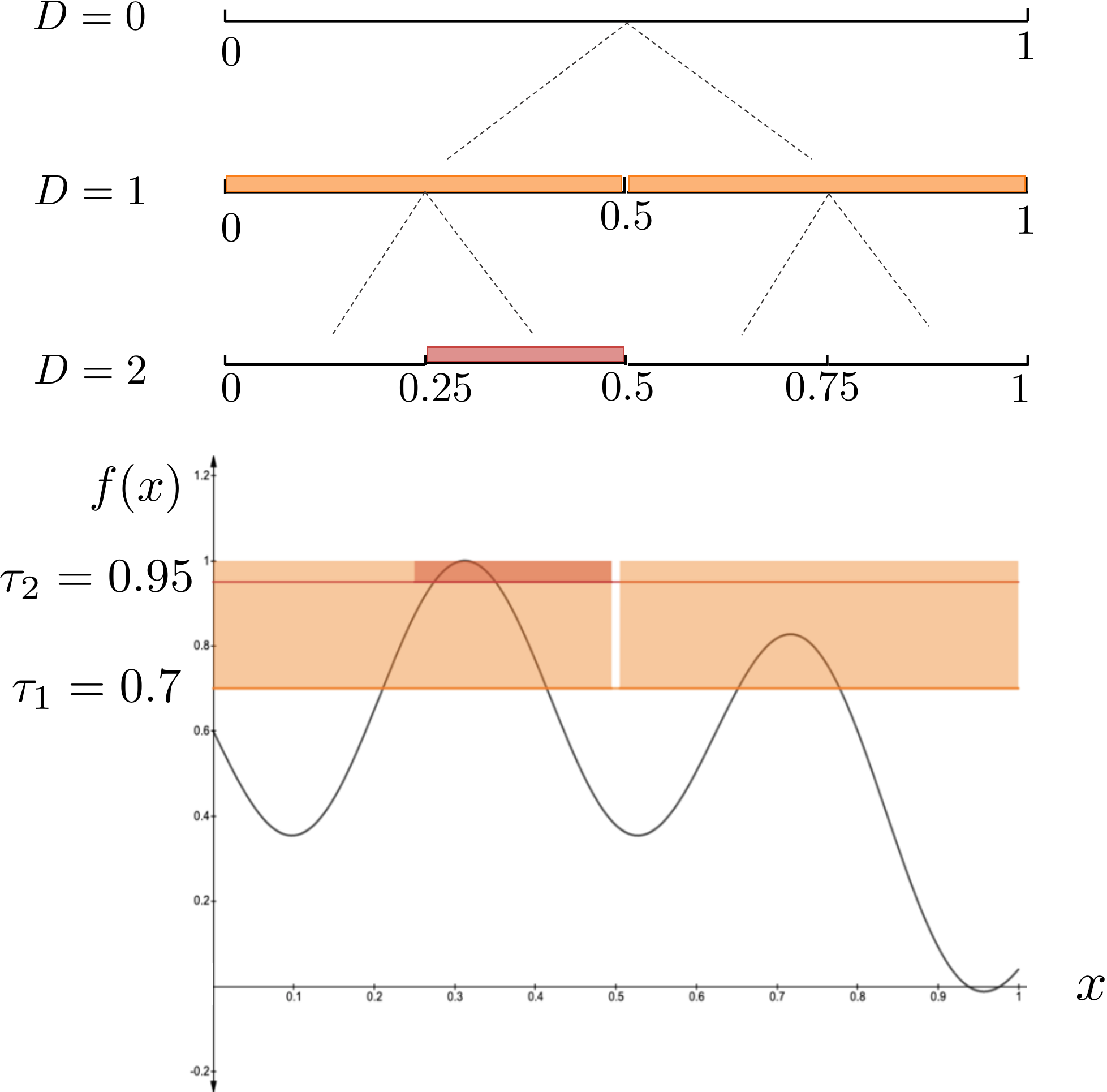}
    \caption{Thresholded domain shrinking.}
    \label{fig:thresholds}
    % \vspace{-1em}
\end{minipage}
~~~~~~~~~~~~~~
\begin{minipage}[h]{0.45\textwidth}
    \centering
    \includegraphics[scale = 0.35]{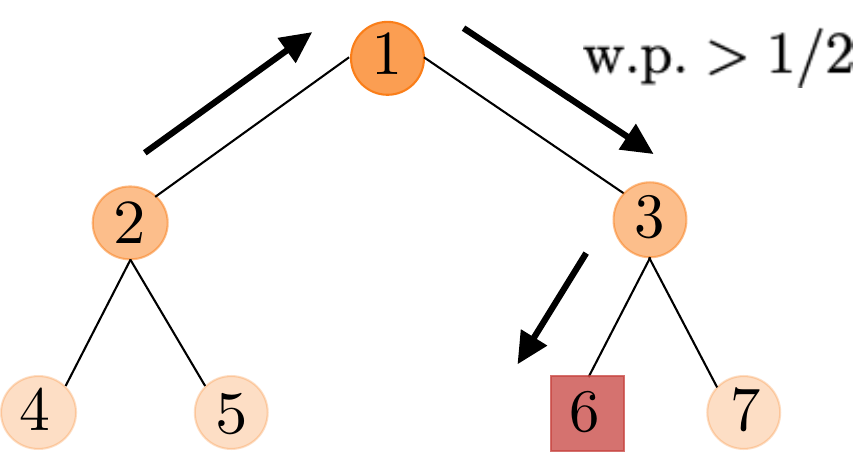}
    \caption{An illustration of the random-walk based search. (Node $6$ is the single high-performing leaf node. If the random walk is currently at node $2$, the correct direction is along the shortest path to node $6$: via node $1$ and then node $3$.)}
    \label{fig:RWT}
    % \vspace{-1em}
\end{minipage}
\end{figure}

% \caption{Thresholded domain shrinking: an example.}
% \caption{A sample path of RWT algorithm. The random walk is initialized at the node indexed $1$ while $6$ denotes the high-performing node. The local test module is first carried out on nodes $2$ and $3$ and based on the results the random walk moves to node $3$ (more likely outcome). Using a similar process, the random walk zooms into node $6$, where it performs the leaf level test and after obtaining a $+1$, the iteration terminates.}

\subsection{Thresholded domain shrinking}
\label{sub:domain_shrinking}

%\paragraph{Tree-based domain shrinking:}
GP-ThreDS operates in epochs. Each epoch completes one cycle of pruning, refining, and threshold updating as detailed below. \textbf{(i) Pruning:} removing sub-performing regions of $\cX$ from future consideration;  \textbf{(ii) Refining:} splitting high-performing regions of $\cX$ into smaller regions for refined search (i.e., zooming in) in future epochs;  \textbf{(iii) Threshold updating:} updating the threshold on function values that defines the criterion for high/sub-performance to be used in the next epoch. The pruning and refining conform to a binary-tree representation of $\cX$ with nodes representing regions of $\cX$ and edges the subset relation (i.e., region splitting). Throughout the paper, we use nodes and regions of $\cX$ interchangeably.  \\

We explain the details with an example. Consider a one-dimensional function over $\cX = [0,1]$ as shown in Fig.~\ref{fig:thresholds}. Assume that it is known $f(x^*) \in [0, 1.4]$. The function threshold $\tau_1$ defining the pruning criterion in the first epoch is set to the mid-point: $\tau_1 = 0.7$. In epoch $1$, the domain $\cX$ is represented by a tree of height $1$ with the root representing the entire domain $[0,1]$ and the two leaf nodes representing the two sub-intervals $[0, 0.5]$ and $(0.5, 1]$ (see Fig.~\ref{fig:thresholds}). In the pruning stage of this epoch, the algorithm determines, with a required confidence, whether each leaf node contains a point with function value exceeding $\tau_1$. Such threshold-exceeding leaf nodes are referred to as high-performing nodes. Otherwise, they are called sub-performing nodes and are pruned, along with their ancestors, from the tree. Suppose that in this example, both sub-intervals $[0, 0.5]$ and $(0.5, 1]$ are identified as high-performing (see Sec.~\ref{sub:random_walk_on_a_tree} on identifying high-performing nodes). Consequently, no node is pruned, and the algorithm proceeds to the refining stage, where each sub-interval splits, and the tree grows to a height of $2$ with four leaf nodes. The threshold is then updated to $\tau_2 = 0.95$ (see below on threshold updating). The increased threshold reflects an adjustment toward a more aggressive pruning in the next epoch as suggested by the presence of (multiple) high-performing nodes in the current epoch. \\

In the second epoch, the pruning stage aims to identify high-performing (defined by $\tau_2$) nodes among the four leaf nodes. Supposed that it is determined leaf node $(0.25,0.5]$ is the only high-performing node. Then the nodes $[0, 0.25], (0.5, 0.75]$ and $(0.75, 1]$ and all their ancestors are pruned. In the refining stage, the high-performing node $(0.25, 0.5]$ splits into two. The threshold is updated to $\tau_3$. The algorithm then progresses into the third epoch, facing the same decision problem on the two leaf nodes (the two children of $(0.25, 0.5]$) of the pruned tree and following the same pruning-refining-threshold updating cycle. \\

For a general $d$-dimensional problem, the basic structure is the same with three simple generalizations. First, the two children of any given node are formed by equally splitting the longest edge of the corresponding $d$-dimensional cuboid (ties broken arbitrarily). Second, in each epoch, the tree grows by $d$ levels ($d=1$ in the above example) in the refining stage by following successive binary splitting $d$ times. The last detail to specify is that if no leaf node is identified as high-performing in an epoch $k$, then the refining stage is bypassed, and the algorithm repeats the search on the same tree (no pruning or refining) with a decreased threshold $\tau_{k+1}$ in the next epoch. The decreased threshold reflects a lowered estimate of $f(x^*)$ based on the absence of high-performing nodes in the current epoch.  \\

The thresholds $\{\tau_k\}_{k\ge 1}$ are updated iteratively using a binary search to approach $f(x^*)$. For each epoch $k$, the algorithm maintains an interval $[a_k, b_k]$ which is believed to contain $f(x^*)$. The threshold $\tau_k$ is set to the mid-point of $[a_k, b_k]$. The initial interval $[a_1, b_1]$ is set to the known range $[a,b]$ of $f(x^*)$. At the end of epoch $k$, if no leaf node is identified as high-performing, we set $a_{k+1} = a_k - (b_k - a_k)/2$ and $b_{k+1} = b_k - (b_k - a_k)/2$, which leads to a decreased threshold in the next epoch. Otherwise, we set $a_{k+1} = \tau_k - c2^{-\alpha \rho_k/d + 1}$ and $b_{k+1} = b_k$, where $\rho_k$ is the height of the tree before the pruning stage of epoch $k$, and $c \in (0,1/2)$ is a hyperparameter (specified in Sec.~\ref{sub:confidence_based_local_test}).  \\

We emphasize that while the proposed domain-shrinking approach conforms to an ever-growing tree, the algorithm can be implemented without storing the entire tree. The only information about the tree that needs to be maintained is the set $\cD_k$ of high-performing leaf nodes identified in the pruning stage of each epoch $k$. A pseudo-code of GP-ThreDS is provided in Appendix~\ref{sec:algo_description}. \\

\subsection{Identifying high-performing nodes} % (fold)
\label{sub:random_walk_on_a_tree}

We now specify the local algorithm for identifying high-performing nodes in a given epoch $k$. Recall that $\cD_{k}$ denotes the set of high-performing nodes identified in epoch $k$. Each node in $\cD_{k}$ has grown $d$ levels and produced $2^d$ leaf nodes in the refining stage of epoch $k$. The objective of epoch $k+1$ is to determine which of the $2^d\, |\cD_{k}|$ newly grown leaves are high-performing nodes defined by~$\tau_{k+1}$.  \\

In epoch $k+1$, the only portion of the tree that is of interest is the $|\cD_{k}|$ subtrees, each of height $d$ with a root in $\cD_{k}$.
Our approach is to treat these subtrees separately, one at a time. We can thus focus on one subtree to describe the algorithm for identifying which of the $2^d$ leaves are high-performing. The terms root, node, and leaf all pertain to this subtree. 
We also omit the epoch index for simplicity.  \\

\subsubsection{A random-walk based search for high-performing nodes}
\label{subsub:RWT}

A straightforward approach to identifying the high-performing nodes is to test each of the $2^d$ leaf nodes directly. This, however, results in a large number of samples at suboptimal points when the dimension $d$ is high. Our approach is inspired by the RWT (Random Walk on a Tree) algorithm recently proposed as a robust and adaptive algorithm for stochastic convex optimization~\cite{Vakili2019a, Vakili2019b, Salgia2020}.  \\

Assume first there is exactly one high-performing node among the $2^d$ leaf nodes. The basic idea is to devise a biased random walk on the tree that initiates at the root and walks towards the high-performing node at the leaf level. As illustrated in Fig.~\ref{fig:RWT} with $d =2$, at a non-leaf node, the random walk can take one of three directions: towards the parent or one of the two children (the parent of the root is itself). The correct direction is to walk along the shortest path to the high-performing leaf node. With the subset relation encoded by the tree, this implies moving to the child containing a threshold-exceeding point or to the parent when neither child contains threshold-exceeding points. Hence, to guide the random walk, a local sequential test is carried out on the two children, one at a time, to determine, at a required confidence level, whether it is threshold-exceeding (see Sec.~\ref{sub:confidence_based_local_test}). The walk then moves to the first child identified as threshold-exceeding (if any) or to the parent otherwise. The confidence level of the local sequential test at each non-leaf node is only required to ensure the walk is correctly biased, i.e., the probability of walking in the correct direction is greater than $1/2$.  \\

On reaching a leaf node, the algorithm enters the \emph{verification} stage to determine whether this node is the high-performing leaf node. If the decision is no, it moves back to the parent of this leaf node, and the random walk resumes. If yes, the algorithm exits (under the assumption of a single high-performing leaf node). This decision can be made by carrying out the same local sequential test that guides the random walk at non-leaf nodes. The only difference is in the required confidence level. Given that a false positive at a leaf cannot be corrected due to exiting while a false negative only resumes the random walk (hence retractable in the future), the confidence level for a positive decision needs to be sufficiently high to ensure the overall regret performance, while a negative decision only needs to ensure the bias of the walk (as in the non-leaf nodes).  \\

When the number of high-performing leaf nodes is unknown and arbitrary in $\{0,1, 2, \dots, 2^d\}$, multiple runs of the random walk are carried out to identify them one by one. In addition, a termination test on the root node is carried out before each run to determine whether there are still unidentified high-performing leaf nodes. See the supplementary for details along with a pseudo code. \\

We emphasize that the local test is carried out using only observations from the current visit to this node; observations from past visits are forgotten. This is to ensure the random-walk nature of the process for tight-analysis. A computational benefit is that the matrices being inverted to compute the posterior distribution are always small, improving the run-time efficiency of the algorithm.

\subsubsection{The local sequential test} % (fold)
\label{sub:confidence_based_local_test}

The last piece of the puzzle in GP-ThreDS is the local sequential test on a given node of a subtree. 
Given a node/region, $D \subseteq \cX$, a threshold $\tau$, and a confidence parameter $\eta \in (0,1)$, the local sequential test needs to determine, with a $1 - \eta$ confidence level, whether $D$ contains a point with function value exceeding $\tau$. \\

The test first builds a discretization of the region $D$, denoted by the set $D_g = \{x_i\}_{i = 1}^{|D_g|}$.
% where $|D_g|$ denotes the size of the set $D_g$. 
The set of points in $D_g$ are chosen to ensure that $\sup_{x \in D} \inf_{y \in D_g} \| x - y \| \leq \Delta$. A simple way to construct such a discretization is to use uniform grids parallel to the axes with a resolution small enough to satisfy the above constraint. \\
%the following:
% \begin{align}
% 	\sup_{x \in D} \inf_{y \in D_g} \| x - y \| \leq \Delta.  
% 	\label{eq:fill_distance}
% \end{align}
The parameter $\Delta$ in epoch $k$ is set to $\Delta_k = (c/L)^{1/\alpha}2^{-\rho_k/d}$ and is used to control the approximation of the function values in $D$. Recall that $L$ is the H{\"o}lder continuity constant while $c \in (0,1/2)$ is a hyperparameter. The local test sequentially queries points in the set $D_g$ to locally estimate $f$. \\

To determine whether there exists a point $x \in D$ with $f(x) \geq \tau$, the test builds a pair of Upper and Lower Confidence Bounds using sequentially drawn samples and compares each of them to prescribed values. If the UCB goes below $\tau - L \Delta^{\alpha}$, indicating that the node is unlikely to contain a $\tau$-exceeding point, the test terminates and outputs a negative outcome. On the other hand, if LCB exceeds $\tau$, then this is a $\tau$-exceeding point with the required confidence level. The test terminates and outputs a positive outcome. If both the UCB and LCB are within their prescribed ``uncertainty" range, the test draws one more sample and repeats the process. A cap is imposed on the total number of samples. Specifically, the test terminates and outputs a positive outcome when the total number of samples exceeds $\bar{S}(p, L \Delta^{\alpha})$.
A description of the test for $s \geq 1$ after being initialized with a point $x_1 \in D_g$ is given in Fig.~\ref{fig:local_test_module}.We would like to emphasize that the posterior mean and variance $\mu_{s-1}$ and $\sigma^{2}_{s-1}$ considered in the description below are constructed only from the samples collected during that particular visit to the current node.

\begin{figure}[!htb]
\begin{center}
\noindent\fbox{
\parbox{0.9\textwidth}{
{

$\bullet$ If $ \max_{x \in D_g} \mu_{s-1}(x) - \beta_s(\eta) \sigma_{s-1}(x) \geq \tau $, terminate and output $+1$.\\ [0.2em]
$\bullet$ If $ \max_{x \in D_g} \mu_{s-1}(x) + \beta_s(\eta) \sigma_{s-1}(x) \leq \tau - L \Delta^{\alpha} $,  terminate and output $-1$.\\ [0.2em]
$\bullet$ Otherwise, query $ x_{s} = \argmax_{x \in D_g} \mu_{s-1}(x) + \beta_s(\frac{\delta_0}{4T}) \sigma_{s-1}(x)$ \\ [0.2em]
$\bullet$ Observe $y_s = f(x_s) + \epsilon_s$ and use~\eqref{eq:posterior_mean} and~\eqref{eq:posterior_variance} to obtain $\mu_s$ and $\sigma_s$. Increment $s$ by $1$. \\ [0.2em]
$\bullet$ Repeat until $s < \bar{S}(\eta, L \Delta^{\alpha})$.\\ If $s = \bar{S}(\eta, L \Delta^{\alpha})$, terminate and output $+1$.
}
}}
\caption{The local sequential test for the decision problem of finding a $\tau$-exceeding point.}
\label{fig:local_test_module}
\end{center}
\vspace{-1em}
\end{figure}

The parameter $\beta_s(\nu):= B + R \sqrt{2(\gamma_{s-1} + 1 + \log(1/\nu))}$ for $\nu \in (0,1)$.
$\gamma_t$ is the maximum information gain at time $t$, defined as $\gamma_t := \max_{A \subset {\cX}: |A| = t} I(y_A; f_A)$.
Here, $I(y_A; f_A)$ denotes the mutual information between $f_A = [f(x)]_{x \in A}$ and $y_A = f_A + \epsilon_A$. Bounds on $\gamma_t$ for several common kernels are known~\cite{Srinivas2012, Vakili2020a} and are sublinear functions of $t$.  \\

The cap $\bar{S}(\eta, L \Delta^{\alpha})$ on the maximum number of samples is given by
\begin{align}
	& \bar{S}(\eta, L \Delta^{\alpha}) = \min \bigg\{t \in \N : \frac{2(1 + 2 \lambda)\beta_t(\eta)|D_g|^{\frac{1}{2}}}{ (L \Delta^{\alpha}) \sqrt{t}}\leq 1  \bigg\} + 1.
	\label{eq:termination_condition}
\end{align}

The cap on the total number of samples prevents the algorithm from wasting too many queries on suboptimal nodes. Without such a cap, the expected number of queries issued by the local test is inversely proportional to $|f(x^*_{D_g}) - \tau|$, where $x^*_{D_g} = \argmax_{x \in D_g} f(x)$. Consequently, small values of $|f(x^*_{D_g}) - \tau|$  would lead to a large number of queries at highly suboptimal points when $f(x^*_{D_g})$ is far from $f(x^*)$. The cap on the number of samples thus helps control the growth of regret at the cost of a potential increase in the approximation error. It also reduces the cost in computing the posterior distribution by limiting the number of queries at a node. \\

Note that when the sequential test reaches the maximum allowable samples and exits with an outcome of $+1$, it is possible that $f(x^*_{D_g}) < \tau$ (i.e., no $\tau$-exceeding points in $D_g$). Thus, $\tau$ may not be a lower bound for the updated belief of $f(x^*)$, as one would expect in the case of an output of $+1$ from the sequential test.
However, using Lemma~\ref{lemma:termination_condition}, we can obtain a high probability lower bound on $\tau - f(x^*_{D_g})$. 
This additional error term is taken into account while updating the threshold as described in Sec.~\ref{sub:domain_shrinking}. The hyperparameter $c$ trades off this error with the size of the discretization. \\

The sequential test can be easily modified to offer asymmetric confidence levels for declaring positive and negative outcomes (as required for in the verification stage of the RWT search) by changing the confidence parameter in $\beta_s$. Details are given in Appendix~\ref{sec:algo_description}. \\

We point out that the construction of the UCB is based on the UCB score employed in IGP-UCB~\cite{Chowdhury2017}. It is straightforward to replace it with other types of UCB scores. The basic thresholded domain shrinking structure of the proposed algorithm is independent of the specific UCB scores, hence generally applicable as a method for improving the computational efficiency and regret performance of GP-UCB family of algorithms. 

% section algorithm_description (end)

%% file: analysis.tex
\section{Performance Analysis} % (fold)
\label{sec:analysis}

In this section, we analyze the regret and computational complexity of GP-ThreDS. Throughout the section, $D \subseteq \cX$ denotes a node visited by GP-ThreDS, $D_g$ denotes its associated discretization, constructed as described in Sec.~\ref{sub:confidence_based_local_test}, and $x^*_{D_g} = \argmax_{x \in D_g} f(x)$.

\subsection{Regret Analysis}
\label{sub:regret_analysis}

The following theorem establishes the regret order of GP-ThreDS. %run with UCB score proposed by~\cite{Chowdhury2017}.

\begin{theorem}
	Consider the GP-ThreDS algorithm as described in Sec.~\ref{sec:algorithm_description}.
	Then, for any $\delta_0 \in (0,1)$, with probability at least $1 - \delta_0$, the regret incurred by the algorithm is given as 
	\begin{align*}
	    R(T) = O(\sqrt{T \gamma_T}\log T(\log T + \sqrt{\log T \log(1/\delta_0)})).
	\end{align*}
% 	$O(\sqrt{T \gamma_T}\log T(\log T + \sqrt{\log T \log(1/\delta_0)}))$.
	\label{thm:regret}
\end{theorem}
    We provide here a sketch of the proof. The regret incurred by GP-ThreDS is analysed by decomposing it into two terms: the regret in the first $k_0$ epochs referred to as $R_1$, and the regret after the completion of the first $k_0$ epochs referred to as $R_2$, where $k_0 = \max\{ k : \rho_k \leq \frac{d}{2\alpha} \log T \}$. To bound $R_1$, we first bound the regret incurred at any node visited during the first $k_0$ epochs using the following decomposition of the instantaneous regret:
    \begin{align*}
        f(x^*) - f(x_t) & =  [f(x^*) - \tau_k + L\Delta_k^{\alpha} ]  +   [ \tau_k - f(x^*_{D_g}) -  L\Delta_k^{\alpha} ] + [ f(x^*_{D_g}) - f(x_t) ].
    \end{align*}
    In the above decomposition, $k$ denotes the epoch index during which the node is visited.
    Each of these three terms are then bounded separately. The third term in the expression is bounded using a similar approach to the analysis of IGP-UCB~\cite{Chowdhury2017} (notice that $x_t$ is the maximizer of the UCB score) that is to bound it by the cumulative standard deviation ($\sum_{s=1}^t\sigma_{s-1}(x_s)$).
    \begin{lemma}
    For any set of sampling points $\{x_1, x_2, \dots, x_t\}$ chosen from $D_g$ (under any choice of algorithm), the following relation holds:
    $ \sum_{s = 1}^t \sigma_{s-1}(x_s) \leq (1 + 2 \lambda)\sqrt{|D_g|t}$, where $\sigma_{s}(x)$ is defined in~\eqref{eq:posterior_variance}.
    \label{lemma:info_gain}
    \end{lemma}
    Since GP-ThreDS ensures a constant-sized discretization at all times (See Lemma~\ref{lemma:constant_nodes}), the above lemma implies that the sum of posterior standard deviations is $O(\sqrt{t})$ resulting in a tight bound corresponding to the third term (that is an $O(\sqrt{\gamma_t})$ tighter than the bound for IGP-UCB which optimizes the UCB score over the entire domain).
    The first two terms are bounded using the following lemma with an appropriate choice of $\Delta_f$.
    \begin{lemma}
	If the local test is terminated by the termination condition at instant $\bar{S}(\delta_2, \Delta_f)$ as defined in~\eqref{eq:termination_condition}, then with probability at least $1 - \delta_2$, we have $\tau - L\Delta^{\alpha} - \Delta_f \leq f(x^*_{D_g}) \leq \tau + \Delta_f$.  
	\label{lemma:termination_condition}
    \end{lemma}

%     individually using following lemmas.
%     % 
%     \begin{lemma}
%     For any set of sampling points $\{x_1, x_2, \dots, x_t\}$ chosen from $D_g$ (under any choice of algorithm), the following relation holds:
%     $ \sum_{s = 1}^t \sigma_{s-1}(x_s) \leq (1 + 2 \lambda)\sqrt{|D_g|t}$, where $\sigma_{s}(x)$ is as defined in~\eqref{eq:posterior_variance}.
%     \end{lemma}
%     % 
%     \begin{lemma}
% 	If the local test is terminated by the termination condition at instant $t_{\term}(\delta_2, \Delta_f)$ as defined in~\eqref{eq:termination_condition}, then with probability at least $1 - \delta_2$, we have $\tau - L\Delta^{\alpha} - \Delta_f \leq f(x^*_{D_g}) \leq \tau + \Delta_f$.  
% 	\label{lemma:termination_condition}
%     \end{lemma}
%     % 
%     \begin{lemma}
% 	Consider the local test module being carried out on a domain $D$, with a threshold $\tau$ and a confidence parameter $p \in (0, 1/2)$ during an epoch $k \geq 1$ of GP-ThreDS. 
% 	If $D$ contains a $\tau$-exceeding point, then the local test module outputs $+1$ with probability at least $1 - p$. If the local test outputs $-1$, then with probability at least $1 - p$, $D$ does not contain a $\tau$-exceeding point.
% 	\label{lemma:local_test}
% 	\end{lemma}
% % 	
% 	\begin{lemma}
% 	Let the interval in which $f(x^*)$ lies, as maintained by the algorithm at the beginning of epoch $k$, be denoted by $[a_k, b_k]$. Then $|b_k - a_k| \leq (1 + 2c\rho_k/d)2^{-\alpha \rho_k/d}$.
% 	\label{lemma:interval_length}
%     \end{lemma}
% % 	
    The final bound on $R_1$ is obtained by a combination of the upper bound on regret on each node and the bound on the total number of nodes visited by GP-ThreDS, captured in the following lemma.
    % which is given by the following lemma. 
    % 
    \begin{lemma}
	Consider the random walk based routine described in Section~\ref{sub:random_walk_on_a_tree} with a local confidence parameter $p \in (0, 1/2)$. Then with probability at least $1 - \delta_1$, one iteration of RWT visits less than $\frac{\log(d/\delta_1)}{2(p - 1/2)^2} $ nodes before termination.
	\label{lemma:number_of_nodes}
    \end{lemma}
    To bound $R_2$, we bound the difference in function values using the H{\"o}lder continuity of the function along with the upper bound on the diameter of the nodes after $k_0$ epochs. Adding the bounds on $R_1$ and $R_2$, we arrive at the theorem. The detailed proofs are provided in Appendix~\ref{sec:proofs}.
    % of the lemmas and the theorem 
    We would like to point out that the regret analysis depends on the choice of the UCB score. While we have used the UCB score of IGP-UCB, this analysis is straightforward to extend to other UCB scores.

% \textcolor{red}{Need to add another theorem/discussion to explicitly state the dependence on dimension $d$. Focus on telling that main work is to reduce the computational complexity wrt length of time horizon and not the dimension. Also since the main focus of the work now is theoretical guarantees on optimal performance, I think it might be better if we first write the theorem about regret guarantees and then about the computational complexity. A reviewer was disappointed that they did not see actual expression for computational savings in the main text, not in $O(\cdot)$ notation.}

\begin{remark}
We note that our assumptions are consistent with those used in proving the lower bounds. In particular, the lower bounds are proven for the Matérn family of kernels including the SE kernel in~\cite{Scarlett2017}.~\cite[Proposition 1]{Shekhar2020} proves the H\"older continuity of this family of kernels. Thus, our assumption on H\"older continuity is consistent with the lower bound. In addition, the proof of lower bound considers a class of functions whose RKHS norm is upper bounded by a known constant~\cite[Sec 1.1]{Scarlett2017}. This upper bound translates to an upper bound on the absolute value of $f$, which is consistent with our assumption on having a finite range for $f$. 
\end{remark}

\subsection{Computational Complexity}
\label{sub:computational_complexity}

% The following theorem establishes a bound on the size of the discretization used in GP-ThreDS.

The following theorem bounds the worst-case overall computational complexity of GP-ThreDS.
\begin{theorem}
    The worst-case overall computational complexity of GP-ThreDS is $O(T^4)$, where $T$ is the time horizon.
    \label{thm:computational_complexity}
\end{theorem}
The proof of theorem follows from the following lemma.
\begin{lemma}
    The number of points in the discretization, $|D_g|$, for any node $D$, is upper bounded by a constant, independent of time. i.e., $|D_g| = O(1)$,  $\forall ~ t \leq T$.
	\label{lemma:constant_nodes}
\end{lemma}
From the lemma, we can conclude that the number of UCB score evaluations in GP-ThreDS is constant at all times $t$, hence matrix inversion becomes the dominant source of computational complexity. Since no more than $t$ samples are used to compute the posterior distribution at time $t$, the worst-case cost associated with matrix inversion step is $O(t^3)$ and consequently the worst-case computational complexity of GP-ThreDS is $O(T^4)$ leading to computational savings of $O(T^{2d - 1})$ over GP-UCB family of algorithms. Lemma~\ref{lemma:constant_nodes} is proven by showing that the rate of domain size shrinking matches the rate of granularity of the discretization across epochs. Thus, the size of discretization does not need to increase with time. 
% For the lemma, the crux of the proof lies in establishing that the rate at which the domain shrinks matches with the rate at which the discretization gets finer across epochs, resulting in a constant sized discretization in each epoch. 
Please refer to Appendix~\ref{sec:proofs} for a detailed proof.  \\

While the discretization does not grow with $t$, it is exponential in $d$. Since non-convex optimization is NP-Hard, such an exponential dependence on $d$ is inevitable for maintaining the optimal learning efficiency. In this work, we focus on reducing the computational complexity with respect to the time horizon~$T$. The proposed domain shrinking technique can be used in conjunction with dimension reduction techniques (e.g.,~\cite{Wang2016}) to achieve efficiency in both $T$ and $d$ (although at the price of invalidating the regret bounds). 

% \vspace{-0.5em}

% The crux of the proof of this theorem lies in establishing that the rate at which the domain shrinks matches with the rate at which the discretization gets finer across epochs. Here we provide a sketch of the proof and defer the detailed proof to the supplementary material. Using bounds on covering numbers, we first show that for a domain $D$ visited during epoch $k$, the size of the discretization, $|D_g|$, is $O(\mathrm{vol}(D) \Delta_k^{-d})$, where $\mathrm{vol}(D)$ is the volume of $D$. Next, we note that during epoch $k$ the domain associated with GP-ThreDS consists of nodes that are at a depth of at least $\rho_k$ on the binary tree. Consequently, we can show that $D$ is a cuboid with edges no longer than $2^{-\rho_k/d}$. Combining this along with the value of $\Delta_k$, we conclude that $|D_g|$ is $O(1)$ for all epochs $k$. 

% From the result in Theorem~\ref{thm:constant_nodes}, it immediately follows that the worst-case overall computational complexity of GP-ThreDS is $O(T^4)$. Since the number of UCB score evaluations in GP-ThreDS is constant at all times $t$, matrix inversion becomes the dominant source of computational complexity. Since no more than $t$ samples are used to compute the posterior distribution at time $t$, the cost associated with matrix inversion step is $O(t^3)$ and consequently the worst-case computational complexity of GP-ThreDS is $O(T^4)$. 

% section analysis (end)

%% file: simulations.tex
\section{Empirical Studies}
\label{sec:simulations}

\begin{figure*}[]
\centering
\subfloat[Branin]{\label{fig:branin_plot}\centering \includegraphics[scale = 0.25]{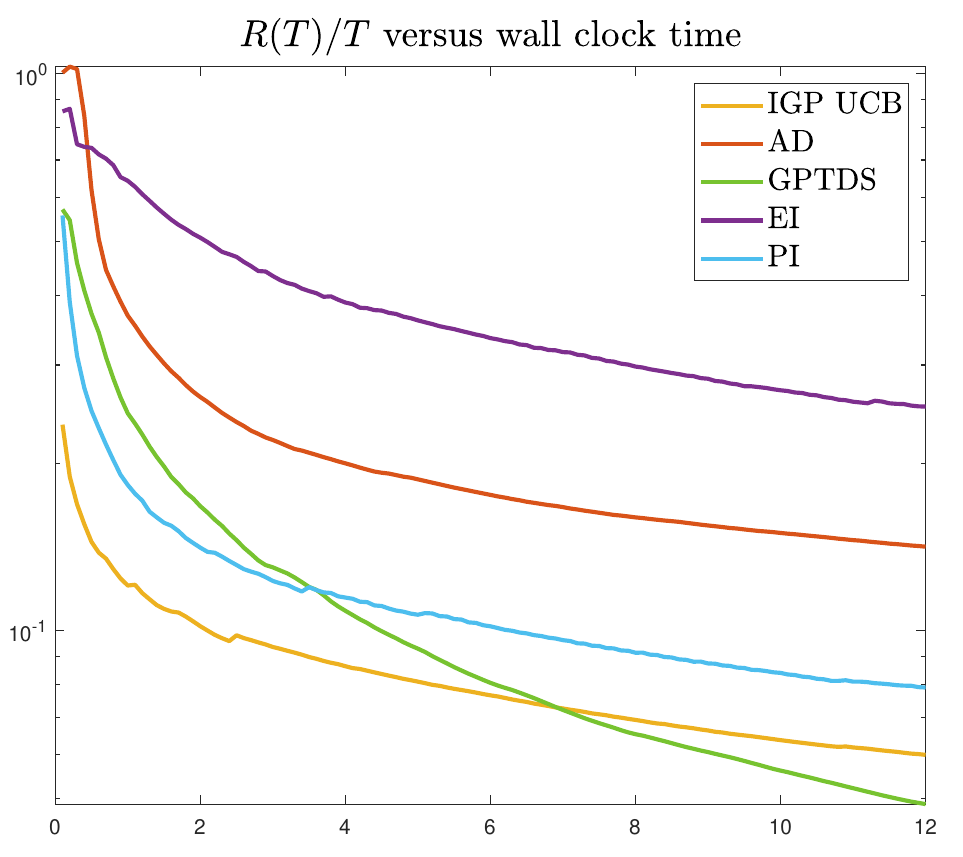}}
~
\subfloat[Rosenbrock]{\label{fig:rosenbrock_plot}\centering \includegraphics[scale = 0.26]{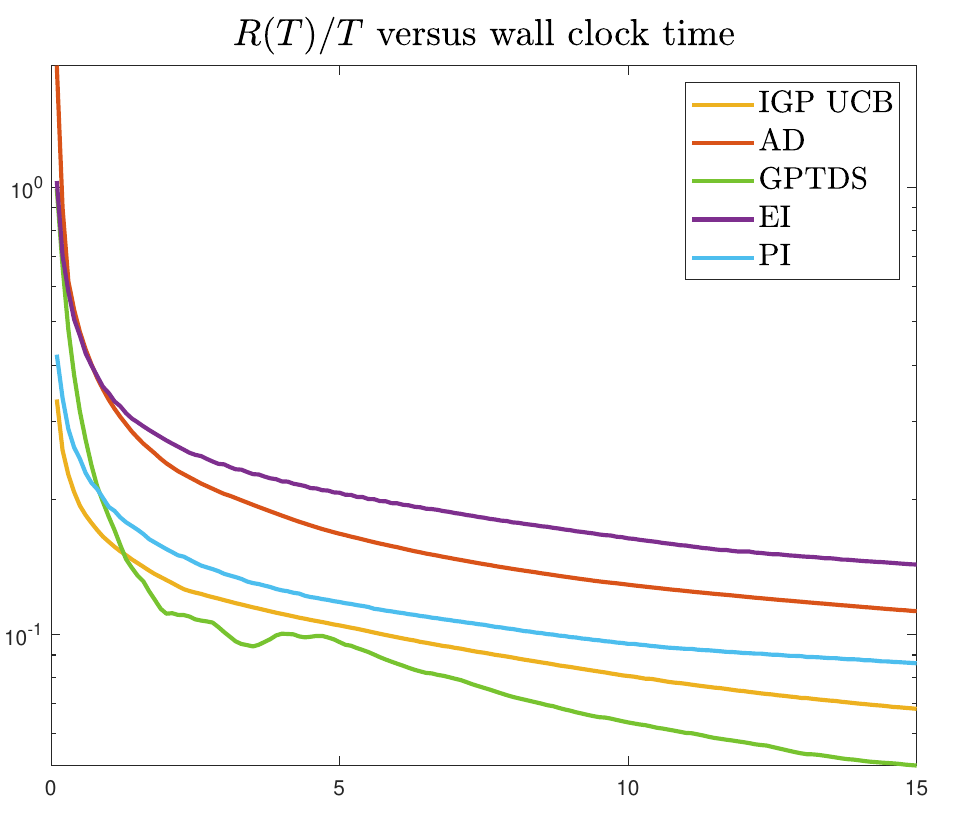}}
~
% \subfloat[CNN]{\label{fig:cnn_plot}\centering \includegraphics[scale = 0.2]{cnn_avg_reg_log_plot_v2.pdf}}
\subfloat[Time taken by diff. algorithms]{\label{fig:time_taken} \centering \includegraphics[scale = 0.25]{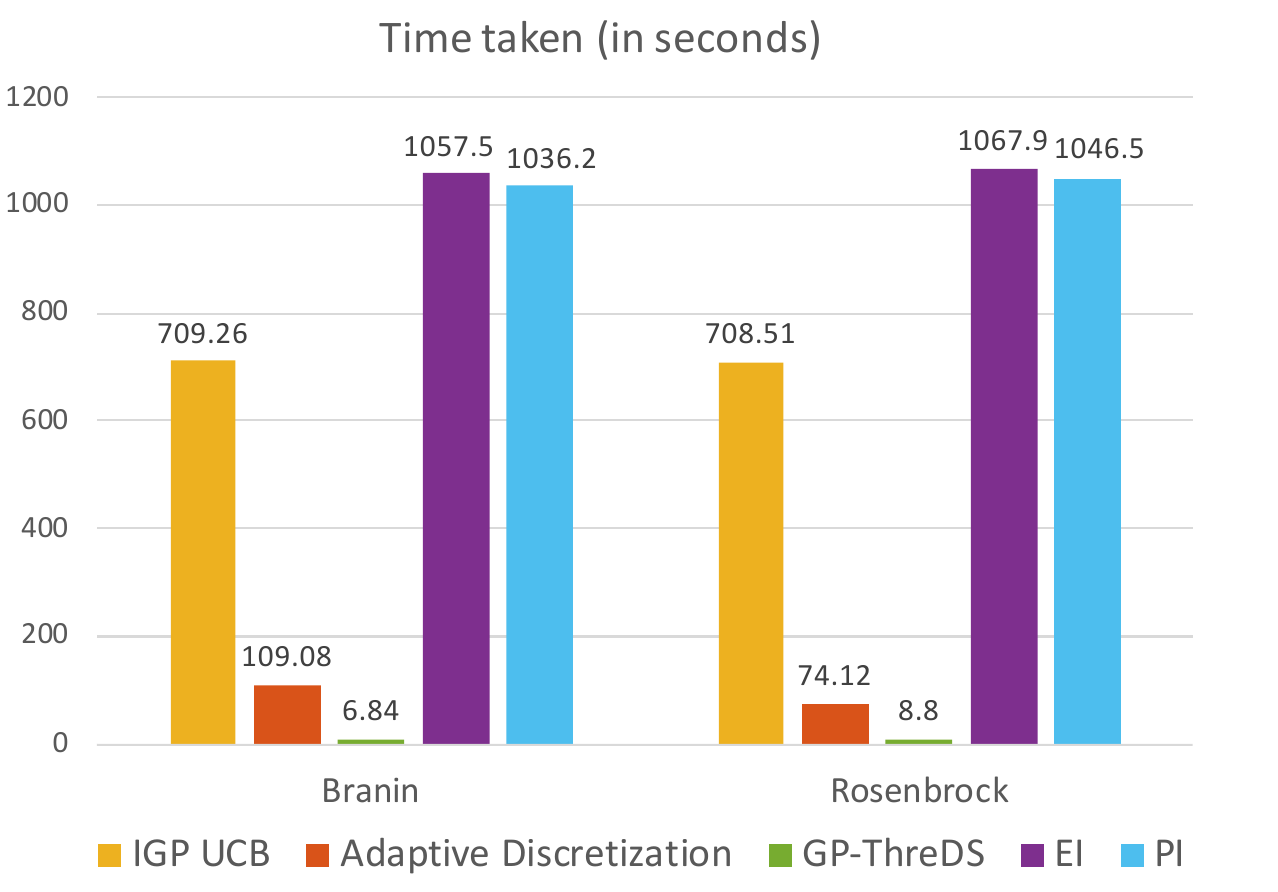}}
\caption{(a)-(b) Average cumulative regret against wall clock time for different algorithms on benchmark functions.  (d) Computation time (in seconds) for $1000$ samples for different algorithms.}
\label{fig:plots}
\vspace{-1em}
\end{figure*}

% and $(0.5, [0.3, 1.4], 0.1)$
In this section, we compare the performance of GP-ThreDS with several commonly used Bayesian optimization algorithms: IGP-UCB~\cite{Chowdhury2017}, Adaptive Discretization (AD)~\cite{Shekhar2018}, Expected Improvement (EI)~\cite{wang2014theoreticalGPEI} and Probability of Improvement (PI)~\cite{wang2018regretGPPI}. For the local test of GP-ThreDS we use the exact same UCB score as the one in IGP-UCB.  \\

We compare these algorithms on two standard benchmark functions
for Bayesian optimization: \emph{Branin} and \emph{Rosenbrock} (see~\cite{Azimi2012, Picheny2013} as well as the supplementary material for their analytical expressions).
We use the SE kernel with lengthscale of $l = 0.2$ on domain $[0,1]^2$. We use a Gaussian noise with variance of $0.01$. The parameters $\lambda$ in the GP model and $R$ in $\beta_t$ are also set to $0.01$. The value of $\delta_0$ is set to  $10^{-3}$. To limit the computational cost in the standard implementation of IGP-UCB, we consider a maximum of $6400$ points in the grid. \\

Figures~\ref{fig:branin_plot},~\ref{fig:rosenbrock_plot}, show the per-sample average regret, in log scale, measured at every $0.1$ seconds, wall clock time. Specifically, within a given time (the X-axis of the figures), different algorithms process different number of samples determined by their computational complexity. The average per sample regret is then shown against the time taken. 
The plots are the average performance over $10$ Monte Carlo runs. As expected from theoretical results, GP-ThreDS achieves the best performance especially as time grows. 
Figure~\ref{fig:time_taken} directly compares the computation time of all algorithms for processing $1000$ samples, averaged over $10$ Monte Carlo runs. GP-ThreDS enjoys a much smaller computation cost in terms of time taken (in seconds).  \\

The details of algorithm parameters, benchmark functions, as well as additional experiments on hyperparameter tuning of a convolutional neural network for image classification are in Appendix~\ref{sec:experiments}.

% section empirical_studies (end)

%% file: conclusion.tex
\section{Conclusion}

A GP-based algorithm witha regret of $\tilde{O}(\sqrt{T\gamma_T})$ for black-box optimization under noisy bandit feedback was proposed. That is order optimal, up to poly-logarithmic factors, for the cases where a lower bound on regret is known. 
% order-optimal regret guarantee \textcolor{red}{(upto poly-logarithmic factors)} for black-box optimization under noisy bandit feedback was proposed. 
The proposed approach is rooted in the methodology of domain shrinking realized through a sequence of tree-based region pruning and refining to concentrate queries in high-performing regions of the function domain. It offers high learning efficiency, allows tight regret analysis, and achieves a computational saving of $O(T^{2d - 1})$ over GP-UCB family of algorithms.

%% file: appendix.tex
\newpage
\appendix

\section*{Appendix}

Additional related works are discussed in Appendix~\ref{sec:related_work}. Further details on the GP-ThreDS algorithm are presented in Appendix~\ref{sec:algo_description}. Proof of Theorem~$1$ and all the lemmas are provided in Appendix~\ref{sec:proofs}. More details on the experiments, as well as additional experiments are given in Appendix~\ref{sec:experiments}.

\section{Additional Related Work}
\label{sec:related_work}

Following the work by \cite{Srinivas2010} on GP-UCB, several extensions have been proposed based on combining GP with bandit techniques.  Representative results include extensions to arbitrary compact metric spaces ~\cite{Contal2016}, contextual bandits~\cite{Krause2011, Valko2013a}, parallel observations~\cite{Desautels2012, Contal2013}, ordinal models~\cite{picheny2019ordinal}, robust optimization~\cite{Bogunovic2018}, and multi-fidelity observations~\cite{Kandasamy2019}. In \cite{Chowdhury2017}, the authors proposed an improved version of GP-UCB
with an improved confidence interval based on a self-normalized concentration inequality that was inspired by similar results derived in~\cite{Abbasi-Yadkori2011} for linear bandits. The query point selection strategy in all these approaches involves optimizing the UCB over the entire domain through an exhaustive search over a grid of $O(t^{2d})$ points at time instant $t$.  \\

There is a growing body of work in the literature addressing the high cost associated with computing the posterior distribution (the first computational bottleneck as discussed in Sec.~\ref{sec:introduction}). Such approaches usually involve approximating the GP posterior by using techniques such as adaptive matrix sketching~\cite{Calandriello2019}, sparse variational inference~\cite{Titsias2009, Hensman2013, Vakili2020a, Huggins2020}, random Fourier features~\cite{Rahimi2009}, linearization~\cite{Kuzborskij2020} and additivity~\cite{Kandasamy2015}.  \\

As pointed out earlier in Sec.~\ref{sec:introduction}, the dominating source of the computational cost is in finding the maximizer of the UCB score. This issue has not received much attention except in a couple of recent studies. Mutn{\'{y}} et al. \cite{Mutny2018} considered a problem where the kernel can be approximated with Quadratic Fourier Features. 
This additional assumption results in a linear model where the UCB proxy can be optimized using an efficient global optimizer. However, this assumption practically limits the GP model to squared exponential kernels. In contrast, the computationally efficient approach proposed in this work is generally applicable.
In \cite{Shekhar2018}, the authors proposed an adaptive discretization approach similar to that of \cite{Bubeck2011b, Wang2014}. The key idea is to replace the uniform discretization in GP-UCB with a non-uniform discretization that adapts to the observed function values so that regions with higher function values enjoy a finer discretization. Nevertheless, the discretization is still carried over the entire function domain throughout the learning process, and a global maximization of the UCB score needs to be carried out at each time instant over a linearly growing set of discrete points. The proposed GP-ThreDS, however, continuously shrinks the function domain and evaluates the UCB score always on a \emph{bounded} set of discrete points. The global maximization objective is also relaxed to determining the existence of threshold-exceeding points.  \\  

Using a tree structure to represent successive partitions of the search domain is a classical approach and has seen its use in the bandit literature~\cite{Bubeck2011b, Munos2011, Kleinberg2008}. Such methods are characterized by growing the tree at nodes with high UCB without pruning the nodes with low values of UCB. This is fundamentally different from the domain shrinking approach of GP-ThreDS. A different tree-based method was considered in \cite{Wang2020} where the tree structure is dynamic and may not be computationally efficient. Furthermore, they did not provide any theoretical results. \\

In contrast to our agnostic regularity assumption on $f$ (being fixed and belonging to an RKHS), a Bayesian setting was also considered in~\cite{Srinivas2010} where $f$ is assumed to be a sample from a GP. The regret bounds were then provided in high probability with respect to both noise and the randomness in $f$. For GP-UCB algorithm,~\cite{Srinivas2010} proved a tighter $O(\sqrt{T\gamma_T})$ bound under the Bayesian setting. Under the Bayesian setting,~\cite{kandasamy2018parallelised} built on ideas
from~\cite{Russo2014,Russo2016Info} to show that GP-TS achieves the same order of regret as
GP-UCB. A Bayesian optimization algorithm for max-value entropy search was shown to enjoy the same regret order as GP-UCB and GP-TS in~\cite{Wang2017}. An $\Omega(\sqrt{T})$ lower bound on regret was proven under the Bayesian setting~\cite{scarlett2018tight, shekhar2021significance} \\

Several works consider a noise-free setting ($\epsilon_t = 0, \forall t$) which results in tighter regret bounds. In particular~\cite{Bull2011} and~\cite{vakili2020noisefree} studied the noise-free Bayesian optimization for Mat{\'e}rn family of kernels, under simple and cumulative regret settings, respectively. The simple regret problem can be addressed using pure exploration algorithms such as epsilon greedy. Under a Bayesian and noise free setting the regret bounds can further improve to exponential rates~\cite{DeFreitas2012, Kawaguchi2015, Grunewalder2010} .  \\

Practitioners' approach to Bayesian optimization generally consists of selecting the observation points based on optimizing the so called acquisition functions (such as UCB in GP-UCB and Thompson sample in GP-TS). Other notable acquisition functions are GP-EI (that stands for expected improvement) and GP-PI (that stands for probability of improvement)~\cite{hoffman2011portfolio}, which are shown to enjoy the same regret guarantees as GP-UCB~\cite{wang2018regretGPPI, nguyen2017regretGPEI, wang2014theoreticalGPEI}. When implementing Bayesian optimization algorithms which are based on an acquisition function (GP-UCB, TS, PI and EI), a practical idea is to use an off-the-shelf optimizer to solve the optimization of the acquisition function at each iteration. This method although can lead to significant gains in computational complexity, invalidates the existing regret bounds. Our focus in this work has been to introduce a practical algorithm with provable regret guarantees.

\section{GP-ThreDS Algorithm}
\label{sec:algo_description}

In this section, we provide a pseudo code for GP-ThreDS as well as additional details on its implementation. 

\subsection{GP-ThreDS Pseudo Code}

A pseudo code for GP-ThreDS is given in Algorithm~\ref{alg:GP_ThreDS} below. In the pseudo-code,\texttt{getHighPerformingNodes} is the routine identifying the high-performing nodes on a tree of depth $d$ that is the routine described in Sec~\ref{sub:random_walk_on_a_tree}. $\cL_{\upsilon}$ denotes the set of high-performing nodes returned by \texttt{getHighPerformingNodes} corresponding to the node $\upsilon$ in $\cD_k$.

\subsection{Random-walk based search for high-performing nodes }
\label{sub:RWT}

We provide additional details of the random walk based search described in Sec.~\ref{subsub:RWT}. We first provide a pseudo code for the random-walk based search strategy for the case of identifying a single high-performing node with confidence level $\delta_{RW}$ in Alg.~\ref{alg:random_walk_on_a_tree}. \\

As the names suggest, the function \texttt{root} returns the root node of the tree, \texttt{parent}, \texttt{leftChild} and \texttt{rightChild} return the parent node, the left child and the right child, respectively, of the node in the argument. \texttt{SequentialTest} is the sequential test routine described in Sec.~\ref{sub:confidence_based_local_test}. In addition to the node and threshold, it takes two confidence parameters as input. If only one is provided, then a sequential test with symmetric confidence levels is carried out as described in Sec.~\ref{sub:confidence_based_local_test}. If two arguments are provided, then the routine carries out the sequential test with asymmetric confidence levels as described in Appendix~\ref{sub:two_sided_local_test}. In this case, the former parameter is considered to be a bound on the probability of a false negative and the latter parameter a bound on the probability of a false positive. Lastly, $p \in (0,1/2)$ is the confidence parameter that is associated with the bias of the random walk, $\hat{\delta}$ is the confidence parameter determined by $\delta_{RW}$, whose exact value is given below.

\begin{algorithm}
    \caption{GP-ThreDS}
    \label{alg:GP_ThreDS}
    \begin{algorithmic}
        \STATE {\bfseries Input:} $\cD_{0} = \{\cX\}$, $[a_1, b_1] = [0,1]$,  $\delta_0 \in (0,1)$ %$p \in (0,1/2)$,
        \STATE \texttt{// The refining stage in epoch $0$ is completed and we have a tree with $2^d$ leaves with root at $\cD_0$.}
        \STATE Set $k \leftarrow 1$, $\rho_1 \leftarrow d$, $\tau_{1} = (a_1 + b_1)2$
        \REPEAT
        \STATE Set $\cD_{k} \leftarrow \emptyset$  %,  $\tau_k = (a_k + b_k)/2
        \STATE \texttt{// Start Pruning Stage}
        \FOR{$\upsilon$ in $\cD_{k-1}$}
        \STATE Set $\cT_{\upsilon}$ to be the tree of depth $d$ rooted at the node $\upsilon \in \cD_{k-1}$
        \STATE $\cL_{\upsilon} \leftarrow$ \texttt{getHighPerformingNodes}($\cT_{\upsilon}$, $\tau_k$, $\delta_0/4T$)
        \STATE $\cD_{k} \leftarrow \cD_{k} \cup \cL_{\upsilon}$
        \ENDFOR
        \IF{$\cD_{k} = \emptyset$}
        \STATE \texttt{// No refining}
        \STATE $\cD_{k} \leftarrow \cD_{k-1}$, $\rho_{k+1} \leftarrow \rho_k$
        \STATE $a_{k+1} \leftarrow a_k - \frac{(b_k - a_k)}{2}$ and $b_{k+1} \leftarrow b_k - \frac{(b_k - a_k)}{2}$
        \ELSE
        \STATE \texttt{//Carry refining stage by growing subtrees rooted at $\upsilon$ for all $\upsilon \in \cD_k$ }
        \STATE $a_{k+1} \leftarrow \tau_k - c2^{-\alpha \rho_k/d + 1}$, $b_{k+1} \leftarrow b_k$, $\rho_{k+1} \leftarrow \rho_k + d$
        \ENDIF
        \STATE \texttt{// Update the threshold}
        \STATE $\tau_{k+1} = (a_{k+1} + b_{k+1})/2$
        \STATE $k \leftarrow k + 1$
        \UNTIL{query budget is exhausted}
    \end{algorithmic}
\end{algorithm}

This strategy can be extended for the case of an unknown number of high-performing leaf nodes as follows. First, each high-performing node is identified using a separate iteration of the routine described in Alg.~\ref{alg:random_walk_on_a_tree}. Thus, several runs of the random walk are carried out. Secondly, in order to address the issue of unknown number of high-performing nodes, a termination test at the root node is carried out to determine whether there is any unidentified high-performing leaf node left. A pseudo code is described in Alg.~\ref{alg:random_walk_on_a_tree_multiple_nodes}.  \\

% One of the major modifications in this version is that once 
In Alg.~\ref{alg:random_walk_on_a_tree_multiple_nodes} a high-performing node is identified, it is not considered in future iterations, in order to avoid redetection. Specifically, while carrying out the sequential test on any node, during the $(r+1)^{\text{th}}$ iteration, on any ancestor of the $r$ identified nodes, the points belonging to the identified high-performing nodes are not considered. This is reflected in updating $\cT$ to $\cT \setminus $ \textsf{retNode}. The parameter $\delta_{RW}$ is set to $\delta_0/4T$ and $\hat{\delta}^{(r)} = \frac{\delta_{0}}{8Tr(r+1)(p - 1/2)^2} \log \left( \frac{4dT}{\delta_{0}}\right)$.   \\

\begin{algorithm}
    \caption{Random-walk based strategy for one high-performing node}
    \label{alg:random_walk_on_a_tree}
    \begin{algorithmic}
        \STATE {\bfseries Input:} Binary tree $\cT$ of depth $d$, threshold $\tau$, confidence level $\delta_{RW}$.
        \STATE Set \textsf{currNode} $\leftarrow$ \texttt{root}($\cT$), \textsf{terminate} $\leftarrow 0$
        \WHILE{\textsf{terminate} $\neq 1$}
        \IF{\texttt{depth}(\textsf{currNode}) $== d$}
        \STATE \textsf{retLeaf} $\leftarrow$ \texttt{SequentialTest}(\textsf{currNode},$\tau, p, \hat{\delta}$)
        \IF{\textsf{retLeaf} $ == 1$}
        \STATE \textsf{terminate} $\leftarrow 1$
        \STATE \textsf{retNode}  $\leftarrow$ \textsf{currNode}
        \ELSE 
        \STATE \textsf{currNode} $\leftarrow$ \texttt{parent}(\textsf{currNode})
        \ENDIF
        \ELSE
        \STATE \textsf{retLeft} $\leftarrow$ \texttt{SequentialTest}(\texttt{leftChild}(\textsf{currNode}),$\tau, p$)
        \IF{\textsf{retLeft} $ == 1$}
        \STATE \textsf{currNode} $\leftarrow$ \texttt{leftChild}(\textsf{currNode})
        \ELSE
        \STATE \textsf{retRight} $\leftarrow$ \texttt{SequentialTest}(\texttt{rightChild}(\textsf{currNode}),$\tau, p$)
        \IF{\textsf{retRight} $ == 1$}
        \STATE \textsf{currNode} $\leftarrow$ \texttt{rightChild}(\textsf{currNode})
        \ELSE
        \STATE \textsf{currNode} $\leftarrow$ \texttt{parent}(\textsf{currNode})
        \ENDIF
        \ENDIF
        \ENDIF
        \ENDWHILE
        \RETURN{\textsf{retNode}}
    \end{algorithmic}
\end{algorithm}

In the experiments, for a simpler implementation, we have devised the search for high-performing nodes at depth $d$, for a given input domain $D$, by directly searching among the leaf nodes. To be complete, a pseudo-code is given in Alg.~\ref{alg:heuristic_RWT} which is slightly different from the pseudo-code given in Alg.~\ref{alg:random_walk_on_a_tree_multiple_nodes} regarding this step.

% \subsection{Heuristic Variant of RWT}

\subsection{Sequential test with asymmetric confidence levels}
\label{sub:two_sided_local_test}

In this section, we describe the sequential test with asymmetric confidence levels. Recall the sequential test with symmetric confidence level $\eta$, given a node/region $D$ and a threshold $\tau$ described in Sec.~\ref{sub:confidence_based_local_test} (see Fig.~\ref{fig:local_test_module}). In the asymmetric case, similarly, we assume a threshold $\tau$ and a node/region $D$ is given. We consider two separate cases based on whether false positive rate is higher than false negative rate or vice versa.  \\

\begin{algorithm}
    \caption{Random-walk based strategy for multiple high-performing node}
    \label{alg:random_walk_on_a_tree_multiple_nodes}
    \begin{algorithmic}
        \STATE {\bfseries Input:} Binary tree $\cT$ of depth $d$, threshold $\tau$, confidence level $\delta_{RW}$.
        \STATE Set \textsf{currNode} $\leftarrow$ \texttt{root}($\cT$), \textsf{terminate} $\leftarrow 0$, $r \leftarrow 1$, \textsf{HPNodes} $\leftarrow \emptyset $, \textsf{retNode} $ = $ NULL
        \WHILE{\textsf{terminate} $\neq 1$}
        \IF{\textsf{currNode} == \texttt{root}($\cT$)}
        \STATE \textsf{retRoot} $\leftarrow$ \texttt{SequentialTest}(\textsf{currNode},$\tau, \hat{\delta}^{(r)}, p$)
        \IF{\textsf{retRoot} $ == -1$}
        \STATE \textsf{terminate} $\leftarrow 1$
        \ENDIF
        \ELSE
        \IF{\texttt{depth}(\textsf{currNode}) $== d$}
        \STATE \textsf{retLeaf} $\leftarrow$ \texttt{SequentialTest}(\textsf{currNode},$\tau, p, \hat{\delta}^{(r)}$)
        \IF{\textsf{retLeaf} $ == 1$}
        \STATE \textsf{retNode}  $\leftarrow$ \textsf{currNode}
        \ELSE 
        \STATE \textsf{currNode} $\leftarrow$ \texttt{parent}(\textsf{currNode})
        \ENDIF
        \ELSE
        \STATE \textsf{retLeft} $\leftarrow$ \texttt{SequentialTest}(\texttt{leftChild}(\textsf{currNode}),$\tau, p$)
        \IF{\textsf{retLeft} $ == 1$}
        \STATE \textsf{currNode} $\leftarrow$ \texttt{leftChild}(\textsf{currNode})
        \ELSE
        \STATE \textsf{retRight} $\leftarrow$ \texttt{SequentialTest}(\texttt{rightChild}(\textsf{currNode}),$\tau, p$)
        \IF{\textsf{retRight} $ == 1$}
        \STATE \textsf{currNode} $\leftarrow$ \texttt{rightChild}(\textsf{currNode})
        \ELSE
        \STATE \textsf{currNode} $\leftarrow$ \texttt{parent}(\textsf{currNode})
        \ENDIF
        \ENDIF
        \ENDIF
        \IF{\textsf{retNode} $!=$ NULL}
        \STATE \textsf{HPNodes} $\leftarrow$ \textsf{HPNodes} $\cup$ \textsf{retNode}
        \STATE $\cT \leftarrow \cT \setminus$ \textsf{retNode}
        \STATE $r \leftarrow r + 1$
        \STATE \textsf{retNode} $ = $ NULL
        \ENDIF
        \ENDIF
        \ENDWHILE
        \RETURN{\textsf{HPNodes}}
    \end{algorithmic}
\end{algorithm}

In the first case, let $1 - \hat{\delta}$ denote the confidence level for declaring whether node $D$ is high-performing ( i.e., it contains a point with function value exceeding $\tau$). Let $1-p$ denote the confidence level for declaring the node is not high-performing (i.e., it does not contain a point with function value exceeding $\tau$). We assume that the false positive rate $\hat{\delta}$ is lower than the false negative rate $p$. In this case the lower confidence bound and the upper confidence bound used in the test will initially be set to $\max_{x \in D_g} \mu_{s-1}(x) - \beta_s(\hat{\delta}) \sigma_{s-1}(x) \geq \tau$ and $\max_{x \in D_g} \mu_{s-1}(x) + \beta_s(p) \sigma_{s-1}(x) \leq \tau-L\Delta^\alpha$, respectively. After $\bar{S}(p)$ steps, the upper confidence bound used in the test will be set to $\max_{x \in D_g} \mu_{s-1}(x) + \beta_s(\hat{\delta}) \sigma_{s-1}(x) \leq \tau-L\Delta^\alpha$. The number of samples is capped to  $\bar{S}(\hat{\delta})$. \\

\begin{algorithm}
    \caption{Heuristic for Updating the domain}
    \label{alg:heuristic_RWT}
    \begin{algorithmic}
        \STATE {\bfseries Input:} Input domain $D$, discretization $D_g$, $\cC$, confidence parameter $\delta$, $x_1 \in D_g$, threshold $\tau$
        \STATE Set \textsf{HPNodes} $\leftarrow \{ \}$ , \textsf{terminate} $\leftarrow 0$, $t \leftarrow 1$, $t_{loc} \leftarrow 1$
        \WHILE{\textsf{terminate} $\neq 1$}
        \IF{$ \displaystyle \max_{x \in D_g} \mu_{t-1}(x) + \beta_t(\delta) \sigma_{t-1}(x) \leq \tau - L \Delta^{\alpha}  $} 
        \STATE \textsf{terminate} $\leftarrow 1$ \hfill {// Stop the search}
        \ELSIF{$ \displaystyle \max_{x \in D_g} \mu_{t-1}(x) - \beta_t(\delta) \sigma_{t-1}(x) \geq \tau $}
        \STATE $c^* \leftarrow \{ c \in \cC : \argmax_{x \in D_g} \mu_{t-1}(x) - \beta_t(\delta) \sigma_{t-1}(x) \in c \} $
        \STATE \textsf{HPNodes} $\leftarrow$ \textsf{HPNodes} $ \cup  \ c^* $ \hfill {// Add the child node to the list of high-performing nodes}
        \STATE $\cC \leftarrow \cC \setminus c^*$ \hfill {// Update the set of children}
        \STATE $D_g \leftarrow D_g \setminus  \{ x : x \in c^* \}$ \hfill {// Update the set of search points}
        \STATE $t_{loc} \leftarrow 0$
        \ELSIF{$t_{loc} == t_{\term}$}
        \STATE $c^* \leftarrow \{ c \in \cC : \argmax_{x \in D_g} \mu_{t-1}(x) - \beta_t(\delta) \sigma_{t-1}(x) \in c \} $ 
        \STATE \textsf{HPNodes} $\leftarrow$ \textsf{HPNodes} $ \cup  \ c^* $
        \STATE $\cC \leftarrow \cC \setminus c^*$ %\hfill{// Equivalent to the termination condition in RWT}
        \STATE $D_g \leftarrow D_g \setminus  \{ x : x \in c^* \}$
        \STATE $t_{loc} \leftarrow 0$
        \ENDIF
        \STATE $x_{t} \leftarrow \argmax_{x \in D_g} \mu_{t-1}(x) + \beta_t(\delta) \sigma_{t-1}(x)$
        \STATE Observe $y_t = f(x_t) + \epsilon_t$
        \STATE Update $t \leftarrow t + 1$, $t_{loc} \leftarrow t_{loc} + 1$ and use the update equations to obtain $\mu_t$ and $\sigma_t$
        \ENDWHILE
        \RETURN{\textsf{HPNodes}}
    \end{algorithmic}
\end{algorithm}

If the sequential test outputs a negative value, we have $f(\xdg) < \tau - L \Delta^{\alpha}$ with probability at least $1-p$, (the probability will be higher, at least $1-\hat{\delta}$, if the sequential test outputs a negative value after $\bar{S}(p)$ steps). If the sequential test outputs a positive value,  we have $f(\xdg) > \tau_k$ with a confidence of $1 - \hat{\delta}$. 
% The test may terminate (after $\bar{S}(\hat{\delta})$ steps) and output a potentially erroneous positive value that is addressed in the analysis by the update rule of the thresholds $\tau_k$ over epochs $k$.
A pseudo code is provided in Alg.~\ref{alg:asymmetric_seq_test_fp}.\\

In the case where $\hat{\delta}$ is the false negative rate and $p$ is false positive rate ($\hat{\delta}<p$), the upper and lower confidence bounds used in the test are set to $\max_{x \in D_g} \mu_{s-1}(x) + \beta_s(\hat{\delta}) \sigma_{s-1}(x) \leq \tau-L\Delta^\alpha$ and $\max_{x \in D_g} \mu_{s-1}(x) - \beta_s(p) \sigma_{s-1}(x) \geq \tau$, respectively. The cap on the number of samples is set to $S(\hat{\delta})$. \\

In this case, if the sequential test outputs a positive value, we have $f(\xdg) > \tau_k$ with probability of at least $1 - p$. If the sequential test outputs a negative value, we have $f(\xdg) < \tau - L \Delta^{\alpha}$ with probability at least $1 - \hat{\delta}$. The cap on the number of samples is set to $S(\hat{\delta})$ which helps us to focus our attention on the high confidence result of $-1$ rather than on the low confidence result of $+1$ resulting from the cap on the number of samples. The test may terminate (after $\bar{S}(\hat{\delta})$ steps) and output a potentially erroneous positive value that is addressed in the analysis by the update rule of the thresholds $\tau_k$ over epochs $k$. A pseudo code is provided in Alg.~\ref{alg:asymmetric_seq_test_fn}.

% \begin{figure}[!htb]
% \begin{center}
% \noindent\fbox{
% \parbox{0.9\textwidth}{
% {

% $\bullet$ If $ \max_{x \in D_g} \mu_{s-1}(x) - \beta_s(\eta) \sigma_{s-1}(x) \geq \tau $, terminate and output $+1$.\\ [0.2em]
% $\bullet$ If $ \max_{x \in D_g} \mu_{s-1}(x) + \beta_s(\eta) \sigma_{s-1}(x) \leq \tau - L \Delta^{\alpha} $,  terminate and output $-1$.\\ [0.2em]
% $\bullet$ Otherwise, query $ x_{s} = \argmax_{x \in D_g} \mu_{s-1}(x) + \beta_s(\frac{\delta_0}{4T}) \sigma_{s-1}(x)$ \\ [0.2em]
% $\bullet$ Observe $y_s = f(x_s) + \epsilon_s$ and use it to obtain $\mu_s$ and $\sigma_s$. Increment $s$ by $1$. \\ [0.2em]
% $\bullet$ Repeat until $s < \bar{S}(\eta, L \Delta^{\alpha})$.\\ If $s = \bar{S}(\eta, L \Delta^{\alpha})$, terminate and output $+1$.
% }
% }}
% \caption{The local sequential test for the decision problem of finding a $\tau$-exceeding point.}
% % \label{fig:local_test_module}
% \end{center}
% \vspace{-1em}
% \end{figure}

\begin{algorithm}[!htb]
    \caption{Sequential Test with Asymmetric Confidence (low false positive rate)}
    \label{alg:asymmetric_seq_test_fp}
    \begin{algorithmic}
        \STATE {\bfseries Input:} Discretization $D_g$, confidence parameters $\hat{\delta}$ and $p$, $x_1 \in D_g$, threshold $\tau$
        \STATE \textsf{terminate} $\leftarrow 0$, $s \leftarrow 1$
        \WHILE{\textsf{terminate} $\neq 1$}
        \IF{$s < \bar{S}(p)$}
        \IF{$ \max_{x \in D_g} \mu_{s-1}(x) - \beta_s(\hat{\delta}) \sigma_{s-1}(x) \geq \tau$}
        \STATE \textsf{terminate} $\leftarrow 1$, \textsf{retVal} $\leftarrow +1$
        \ELSIF{$ \max_{x \in D_g} \mu_{s-1}(x) + \beta_s(p) \sigma_{s-1}(x) \leq \tau - L \Delta^{\alpha} $}
        \STATE \textsf{terminate} $\leftarrow 1$, \textsf{retVal} $\leftarrow -1$
        \ELSE
        \STATE Query $ x_{s} = \argmax_{x \in D_g} \mu_{s-1}(x) + \beta_s(\frac{\delta_0}{4T}) \sigma_{s-1}(x)$
        \STATE  Observe $y_s = f(x_s) + \epsilon_s$ and use it to obtain $\mu_s$ and $\sigma_s$
        \STATE $s \leftarrow s + 1$
        \ENDIF
        \ENDIF
        \IF{$s \geq \bar{S}(p)$ and $s < \bar{S}(\hat{\delta})$}
        \IF{$ \max_{x \in D_g} \mu_{s-1}(x) - \beta_s(\hat{\delta}) \sigma_{s-1}(x) \geq \tau$}
        \STATE \textsf{terminate} $\leftarrow 1$, \textsf{retVal} $\leftarrow +1$
        \ELSIF{$ \max_{x \in D_g} \mu_{s-1}(x) + \beta_s(\hat{\delta}) \sigma_{s-1}(x) \leq \tau - L \Delta^{\alpha} $}
        \STATE \textsf{terminate} $\leftarrow 1$, \textsf{retVal} $\leftarrow -1$
        \ELSE
        \STATE Query $ x_{s} = \argmax_{x \in D_g} \mu_{s-1}(x) + \beta_s(\frac{\delta_0}{4T}) \sigma_{s-1}(x)$
        \STATE  Observe $y_s = f(x_s) + \epsilon_s$ and use it to obtain $\mu_s$ and $\sigma_s$
        \STATE $s \leftarrow s + 1$
        \ENDIF
        \ENDIF
        \IF{$s == \bar{S}(\hat{\delta})$ }
        \STATE \textsf{terminate} $\leftarrow 1$, \textsf{retVal} $\leftarrow +1$
        \ENDIF
        \ENDWHILE
        \RETURN \textsf{retVal}
    \end{algorithmic}
\end{algorithm}

\begin{algorithm}[!htb]
    \caption{Sequential Test with Asymmetric Confidence (low false negative rate)}
    \label{alg:asymmetric_seq_test_fn}
    \begin{algorithmic}
        \STATE {\bfseries Input:} Discretization $D_g$, confidence parameters $\hat{\delta}$ and $p$, $x_1 \in D_g$, threshold $\tau$
        \STATE \textsf{terminate} $\leftarrow 0$, $s \leftarrow 1$
        \WHILE{\textsf{terminate} $\neq 1$}
        \IF{$s < \bar{S}(\hat{\delta})$}
        \IF{$ \max_{x \in D_g} \mu_{s-1}(x) - \beta_s(p) \sigma_{s-1}(x) \geq \tau$}
        \STATE \textsf{terminate} $\leftarrow 1$, \textsf{retVal} $\leftarrow +1$
        \ELSIF{$ \max_{x \in D_g} \mu_{s-1}(x) + \beta_s(\hat{\delta}) \sigma_{s-1}(x) \leq \tau - L \Delta^{\alpha} $}
        \STATE \textsf{terminate} $\leftarrow 1$, \textsf{retVal} $\leftarrow -1$
        \ELSE
        \STATE Query $ x_{s} = \argmax_{x \in D_g} \mu_{s-1}(x) + \beta_s(\frac{\delta_0}{4T}) \sigma_{s-1}(x)$
        \STATE  Observe $y_s = f(x_s) + \epsilon_s$ and use it to obtain $\mu_s$ and $\sigma_s$
        \STATE $s \leftarrow s + 1$
        \ENDIF
        \ENDIF
        \IF{$s == \bar{S}(\hat{\delta})$ }
        \STATE \textsf{terminate} $\leftarrow 1$, \textsf{retVal} $\leftarrow +1$
        \ENDIF
        \ENDWHILE
        \RETURN \textsf{retVal}
    \end{algorithmic}
\end{algorithm}

\input{proofs}

\section{Supplemental Material on Experiments}
\label{sec:experiments}

In this section, we provide further details on the experiments, as well as additional experiments on a hyperparameter tuning problem. 

\subsection{Details of Benchmark Functions, Algorithms and their Parameters}
\label{sub:expt_algo_description}

We used two standard benchmark functions, Branin and Rosenbrock in our experiments. The analytical expression for these functions is given below~\cite{Azimi2012, Picheny2013}
\begin{itemize}
    \item \emph{Branin}: $\displaystyle f(x, y) = -\frac{1}{51.95}\left( \left( v - \frac{5.1u^2}{4 \pi^2} + \frac{5u}{\pi} - 6\right)^2 + \left(10 - \frac{10}{8\pi} \right)\cos(u) - 44.81\right)$, where $u = 15x - 5$ and $v = 15y$.
    \item \emph{Rosenbrock}: $f(x, y) = 10 - 100(v-u)^2 - (1 - u)^2$, where $u = 0.3x + 0.8$ and $v = 0.3y + 0.8$.
\end{itemize}

The domain is set to $\mathcal{X} = [0,1]^2$. The details of IGP-UCB, AD, PI and EI are provided below. 

\begin{enumerate}
    \item IGP-UCB: The algorithm is implemented exactly as outlined in~\cite{Chowdhury2017} with $B$ (in scaling parameter $\beta_t$) set to $0.5$ and $2$ for Branin and Rosenbrock, respectively. The parameters $R$ and $\delta_0$ are set to $10^{-2}$ and $10^{-3}$ in both experiments. $\gamma_t$ was set to $\log t$. The size of discretization is increased over time, starting from $400$ points at the beginning and capped at $6400$ points.
    \item Adapative Discretization (AD): The algorithm and its parameters are implemented exactly as described in~\cite{Shekhar2018}. 
    \item Expected Improvement(EI)/Probability of Improvement (PI): Similar to IGP-UCB, EI and PI select the observation points based on maximizing an index often referred to as an acquisition function. The acquisition function of EI is $(\mu(x) - f^* - \varepsilon)\Phi\left( z \right) + \sigma(x)\phi(z)$, where $z = \frac{\mu(x) - f^* - \varepsilon}{\sigma(x)}$. The acquisition function of PI is $\Phi(z)$, where $z = \frac{\mu(x) - f^* - \xi}{\sigma(x)}$. Here, $\Phi(\cdot)$ and $\phi(\cdot)$ denote the CDF and PDF of a standard normal random variable. $f^*$ is set to be the maximum value of $\mu(x)$ among the current observation points. The parameters $\varepsilon$ and $\xi$ are used to balance the exploration-exploitation trade-off. We follow~\cite{hoffman2011portfolio} that showed the best choice of these parameters are small non-zero values. In particular, $\varepsilon$ and $\xi$ are both set to $0.01$.
    % can be used to ensure a certain increase in the acquisition function in each iteration. While in principle they can be set to $0$, we noticed that a small non-zero value resulted in improved performance of these algorithms. Thus, we set $\varepsilon = 0.01 = \xi$.
    \item GP-ThreDS: A pseudo-code is given in Alg.~\ref{alg:heuristic_RWT}. The parameter $\beta_t$ is set exactly in the same way as in IGP-UCB. The initial interval $[a,b]$ is set to $[0.5, 1.2]$ for Branin and $[3, 12]$ for Rosenbrock. The parameter $c$ is set to $0.2$ for both functions.
\end{enumerate}

\begin{figure}[!htb]
    \centering
    \includegraphics[scale = 0.45]{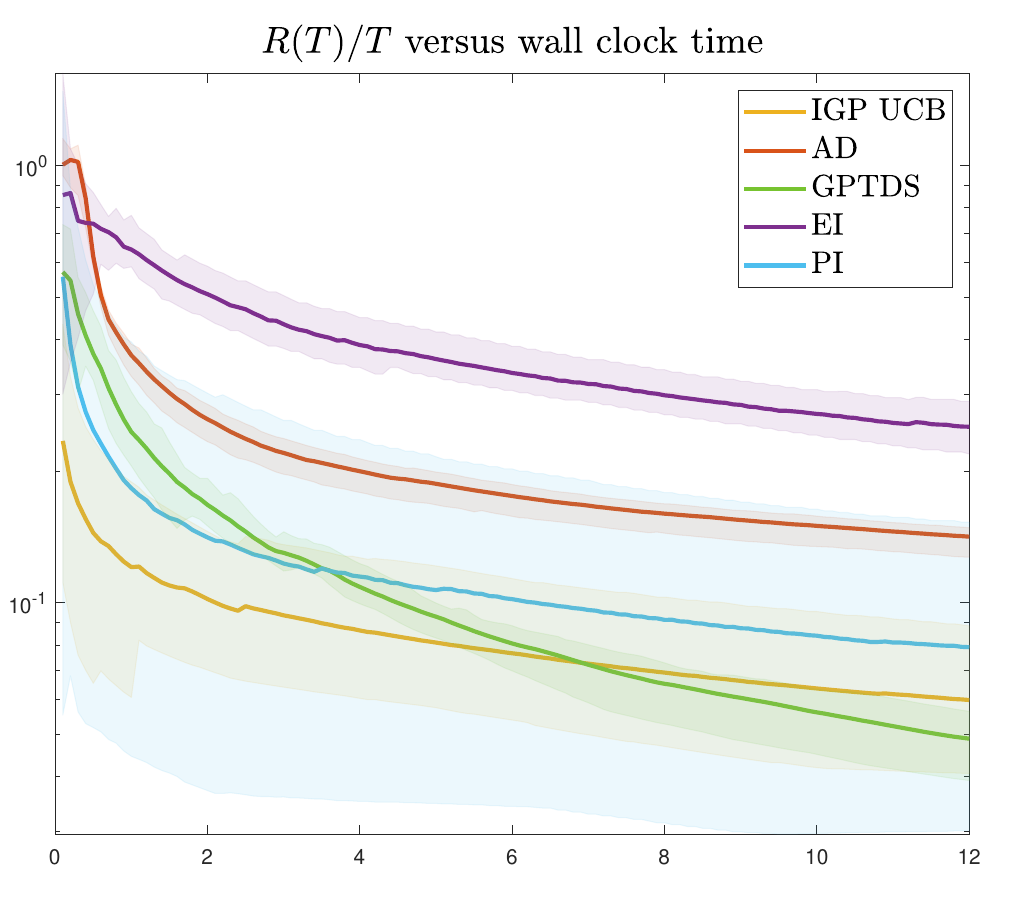}
    \caption{Average regret against wall clock time for Branin}
    \label{fig:branin_err_bars}
\end{figure}

\subsection{Additional Experiments}
\label{sub:add_graphs}

\begin{figure}
\subfloat[]{\label{fig:rosenbrock_err_bars_1}\centering \includegraphics[scale = 0.45]{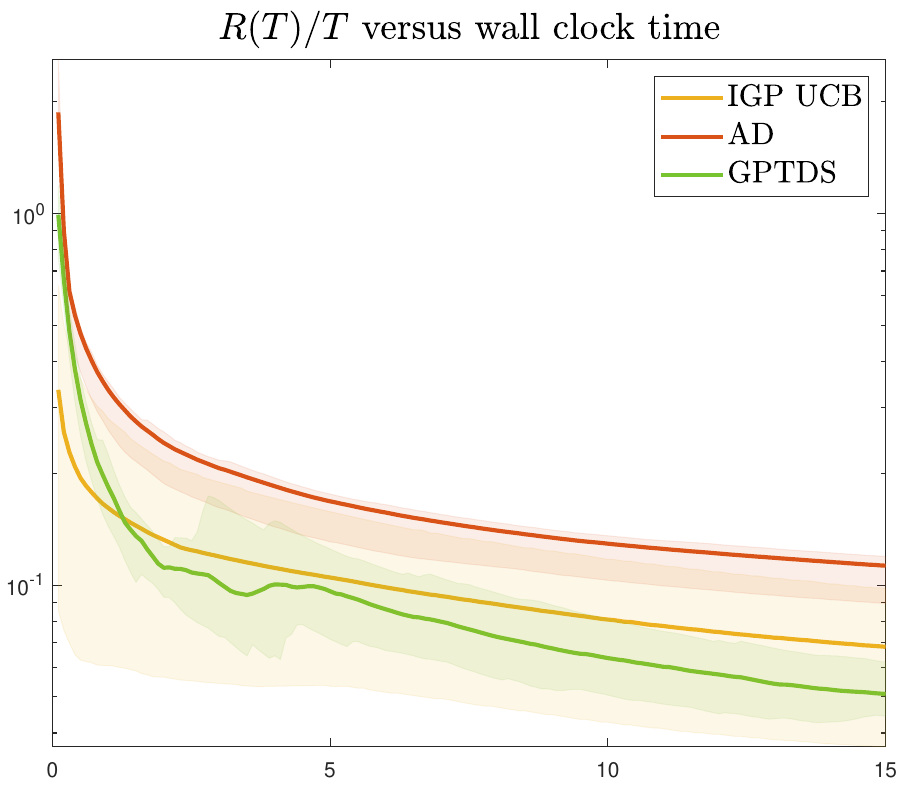}}
~~~~~~~~~~~~
\subfloat[]{\label{fig:rosenbrock_err_bars_2}\centering \includegraphics[scale = 0.45]{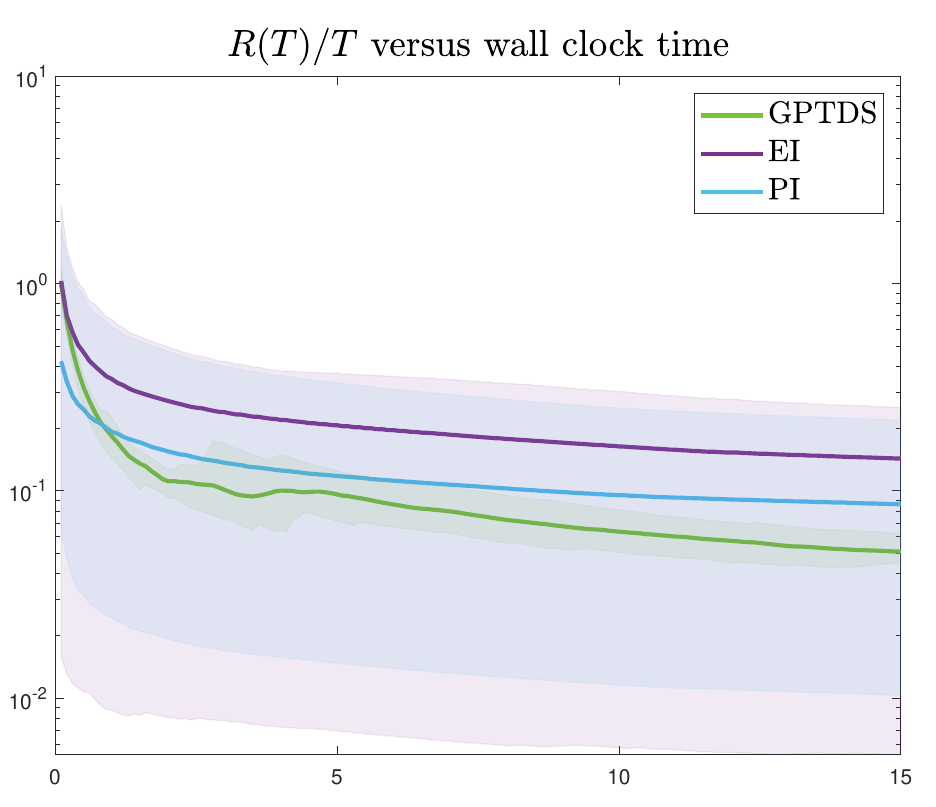}}
\caption{Average regret against wall clock time for Rosenbrock}
    \label{fig:rosenbrock_err_bars}
\end{figure}

We have replotted the average regret against wall clock time for different algorithms on benchmark functions (the same as Fig.~\ref{fig:plots}), with error bars in Fig.~\ref{fig:branin_err_bars} and Fig.~\ref{fig:rosenbrock_err_bars}.

\subsection{Hyperparameter Tuning for a convolutional neural network}
\label{sub:cnn_hyperparameter_tuning}

In this section, we provide additional experiments on using the Bayesian optimization algorithms for hyperparameter tuning for a convolutional neural network (CNN) on an image classification task.  
We have considered the task of digit classification on the MNIST dataset. For the experiments, we have considered a smaller training dataset that contains only $12000$ images instead of $50000$ images in the original dataset. This smaller training dataset is created by randomly sampling $1200$ images corresponding to each digit, making a total of $12000$ images. This dataset is split to training and validation sets of size $10000$ and $2000$, respectively. The split is done in a way that each label has equal numbers of images in the training and the validation set. We used the same test set of $10000$ images as in the original MNIST data set. \\

We consider a simple CNN with 2 convolutional layers followed by two fully connected feedforward layers. We use the ReLU activation function and a max pooling layer with stride $2$ after each convolutional layer. The performance of the algorithms is evaluated on the task of tuning the following five hyperparameters of this CNN.
\begin{itemize}
    \item Batch size: We considered 8 possible values of the batch sizes given by $\{2^3, 2^4, \dots, 2^{10}\}$.
    \item Kernel size of the first convolutional layer with possible values in $\{3,5,7,9\}$.
    \item Kernel size of the second convolutional layer with possible values in $\{3,5,7,9\}$.
    \item Number of (hidden) nodes in the first feedforward layer: The possible values for this hyperparameter belonged to $\{10, 11, 12, \dots, 38, 39, 40\}$.
    \item Initial learning rate: We used stochastic gradient descent with momentum to optimize the loss function. This parameter defined the initial learning rate for the optimizer and it took values in $\{10^{-6}, 10^{-5}, \dots, 10^{-1} \}$.
\end{itemize}

\begin{figure}[!htb]
\centering
\subfloat[Average regret against wall clock time for CNN.]{\label{fig:cnn_err_bars}\centering \includegraphics[scale = 0.42]{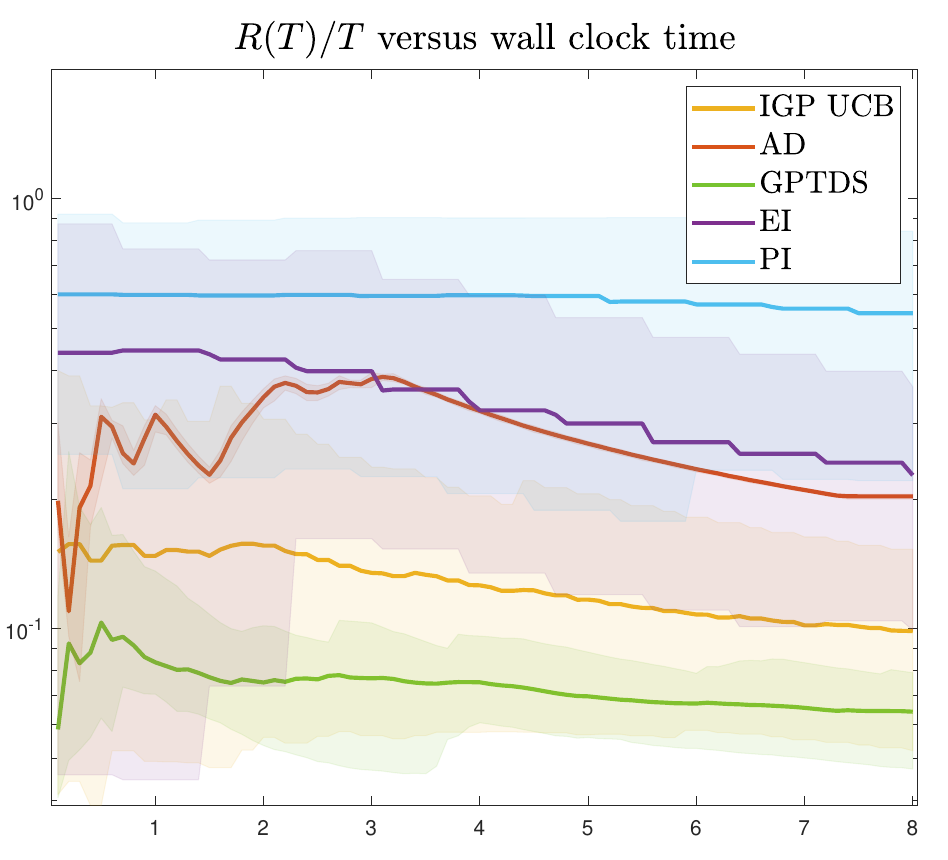}}
~~~~~~
\subfloat[Average regret against wall clock time for EI and PI with a longer time horizon]{\label{fig:cnn_ei_pi}\centering \includegraphics[scale = 0.45]{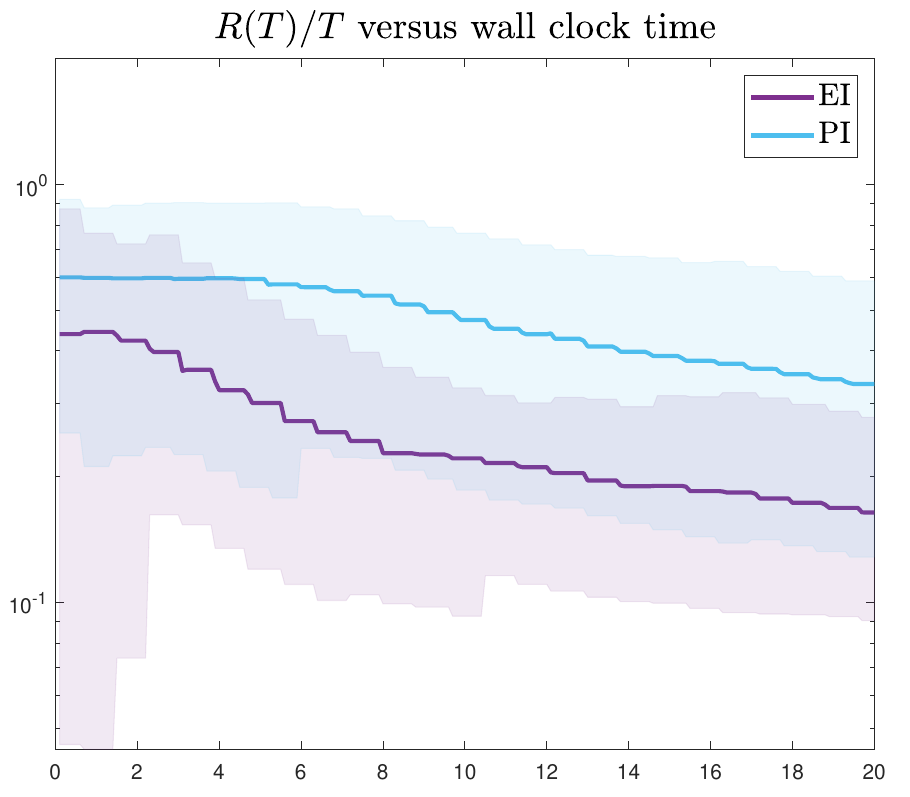}}
\\
\subfloat[Optimization time spent by different algorithms for $T = 50$ samples]{\label{fig:cnn_time}\centering \includegraphics[scale = 0.36]{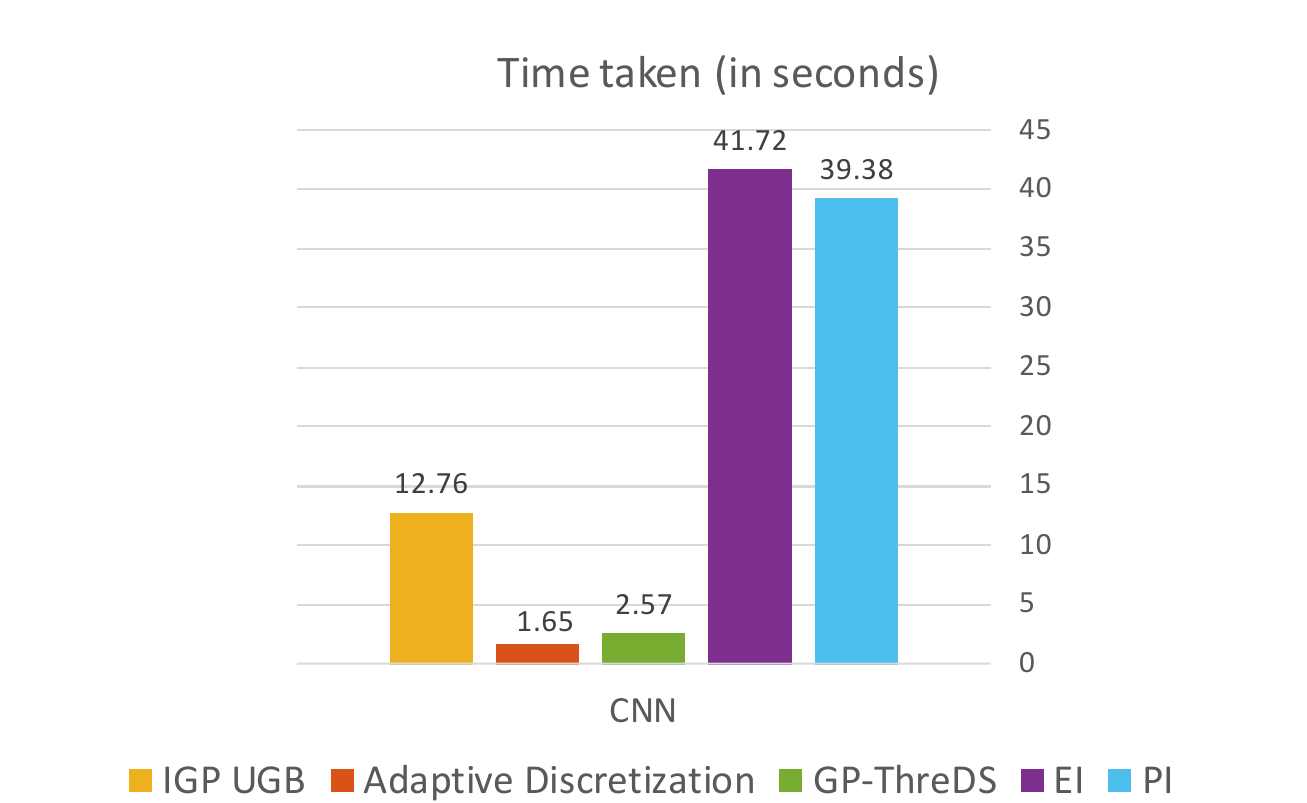}}
~~~
\subfloat[Total time spent by different algorithms for $T = 50$ samples]{\label{fig:cnn_time_total}\centering \includegraphics[scale = 0.36]{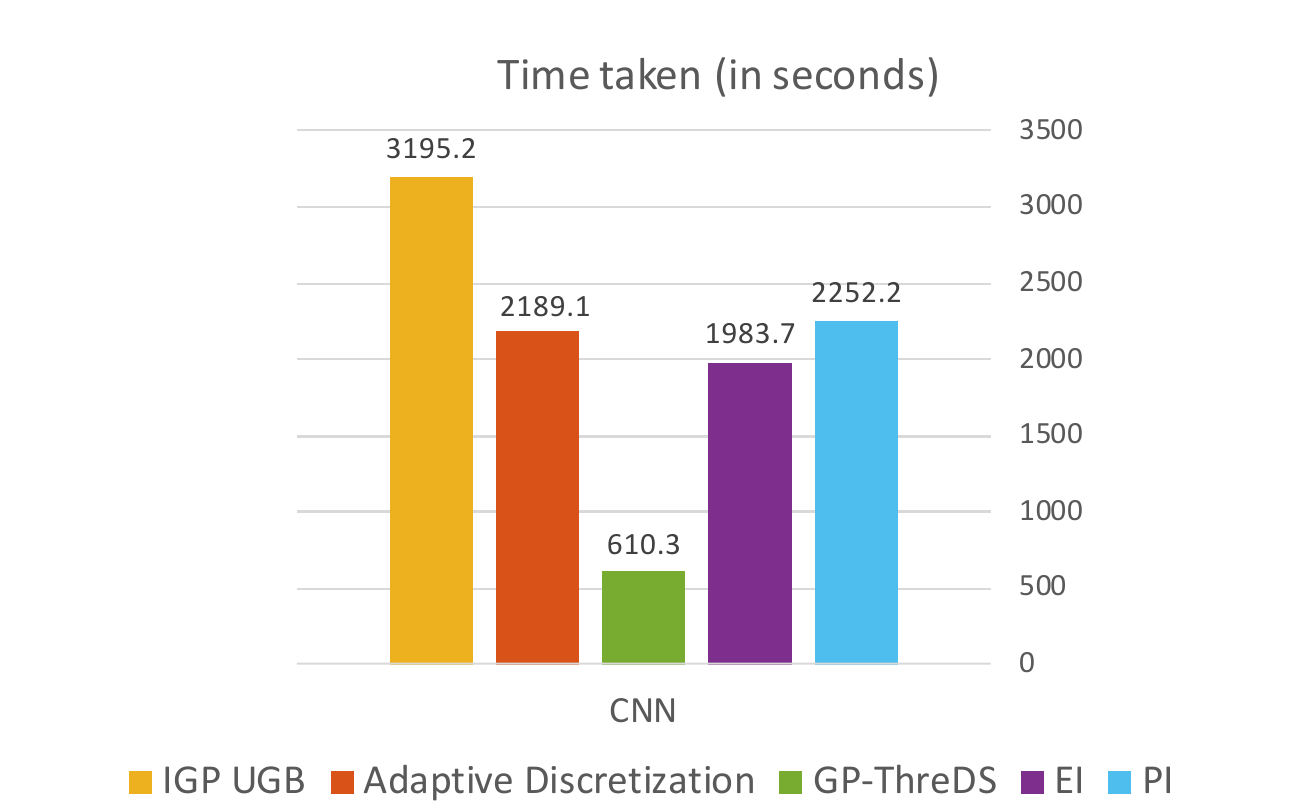}}
\caption{Performance of different algorithms for tuning the hyperparameters of a CNN for image classification.}
\end{figure}

In the implementation, all these parameters were mapped to $[0,1]$ with distinct intervals corresponding to each discrete value. The kernel sizes and the number of hidden nodes were mapped linearly to the interval while the other two parameters were mapped on a log scale, that is, $\log_2($batch-size$)$ and $\log_{10}($learning-rate$)$ were mapped uniformly to the interval $[0,1]$. \\

For this task, the underlying function was modelled using a Mat\'ern kernel with smoothness parameter $2.5$ and lengthscale $l = 0.2$. For this kernel $\gamma_t$ was set to $\sqrt{t}$ and the noise variance was set to $0.0001$. The implementation of all the algorithms is similar to the description in Sec.~\ref{sub:expt_algo_description}.  \\

For the exploration parameter $\beta_t$, $B$ and $R$ are set to $0.5$ and $10^{-4}$, respectively, for both IGP-UCB and GP-ThreDS. The confidence parameter $\delta_0$ is set to $2.5 \times 10^{-2}$ for IGP-UCB and $2 \times 10^{-2}$ for GP-ThreDS. The slightly higher confidence for GP-ThreDS is chosen because it runs for a longer horizon to achieve the same compute time. In GP-ThreDS, the interval $[a,b]$ is set to $[0.3, 1.4]$ and $c$ is set to $0.1$. \\

All the computations were carried out using MATLAB 2019a on a computer with 12 GB RAM and Intel i7 processor (3.4 GHz) with an overall compute time of around 200 hours.  \\

The results for average regret against wall clock time, averaged over $10$ Monte Carlo runs, are shown in Fig.~\ref{fig:cnn_err_bars}. As it can be seen from the figure, GP-ThreDS offers a better performance compared to all other algorithms. It is important to point out that the time on $X$-axis does not include the time spent to train the CNN, it only includes the time spent on Bayesian optimization algorithms assuming the values of the objective function $f$ (here the performance of CNN) are accessible in zero time. To be clear we refer to this time as \emph{Optimization time}. The optimization time spent by different algorithms for $T = 50$ is shown in Fig.~\ref{fig:cnn_time}. We also report the corresponding \emph{total time} contrasting the optimization time that also includes the time spent to train the CNN, in Fig.~\ref{fig:cnn_time_total}. It can be seen that the total time for GP-ThreDS is significantly lower than all other algorithms. Its optimization time however is comparable to AD, while significantly lower than IGP-UCB, EI, PI. It seems that, due to its exploration scheme, AD selects hyperparameters which lead to longer training times. PI and EI in comparison seem to take longer to converge as shown in Fig.~\ref{fig:cnn_ei_pi}.

% For AD, while the actual compute time is comparable to GP-ThreDS, we believe that the overall runtime is significantly larger because of the sampling scheme it employs. Since it explores through fixed points on the tree, it often runs into set of parameters that lead to larger training times leading in overall increased actual runtime. Similarly, for GP-EI and GP-PI, we believe that due to initial variance, we begin to observe the convergence only when plotted for a longer period of time as shown in Fig.~\ref{fig:cnn_ei_pi}.

%% file: proofs.tex
\newtheorem{innercustomthm}{Lemma}
\newenvironment{customlemma}[1]
  {\renewcommand\theinnercustomthm{#1}\innercustomthm}
  {\endinnercustomthm}

\section{Detailed Proofs}
\label{sec:proofs}

In this section, we provide the proof of Theorem{}~\ref{thm:regret}, as well as all the lemmas that are used in the proof of theorems. We first state all the lemmas. We then provide the proof of Theorem~\ref{thm:regret} followed by the proof of the lemmas.

\begin{customlemma}{1}
For any set of sampling points $\{x_1, x_2, \dots, x_t\}$ chosen from $D_g$ (under any choice of algorithm), the following holds:
$ \sum_{s = 1}^t \sigma_{s-1}(x_s) \leq (1 + 2 \lambda)\sqrt{|D_g|t}$. %where $\sigma_{s}(x)$ is defined in~\eqref{eq:posterior_variance}.
% \label{lemma:info_gain}
\end{customlemma}
\begin{customlemma}{2}
If the local test is terminated by the termination condition at instant $\bar{S}(\delta_2, \Delta_f)$ defined as $\bar{S}(\delta_2, \Delta_f) = \min \left\{t \in \N : {2(1 + 2 \lambda)\beta_t(\delta_2)} \sqrt{\frac{|D_g|}{t}}\leq \Delta_f   \right\} + 1$, then with probability at least $1 - \delta_2$, we have $\tau - L\Delta^{\alpha} - \Delta_f \leq f(x^*_{D_g}) \leq \tau + \Delta_f$.  
% \label{lemma:termination_condition}
\end{customlemma}
\begin{customlemma}{3}
Consider the random walk based routine described in Section~\ref{sub:RWT} with a local confidence parameter $p \in (0, 1/2)$. Then with probability at least $1 - \delta_1$, one iteration of RWT visits less than $\frac{\log(d/\delta_1)}{2(p - 1/2)^2} $ nodes before termination.
% \label{lemma:number_of_nodes}
\end{customlemma}
\begin{customlemma}{4}
The number of points in the discretization, $|D_g|$, for any node $D$, is upper bounded by a constant, independent of time. i.e., $|D_g| = O(1)$,  $\forall ~ t \leq T$.
% \label{lemma:constant_nodes}
\end{customlemma}
\begin{lemma}
Consider the local test module carried out on a domain $D$, with a threshold $\tau$ and a confidence parameter $p \in (0, 1/2)$, during an epoch $k \geq 1$ of GP-ThreDS. 
If $D$ contains a $\tau$-exceeding point, then the local test module outputs $+1$ with probability at least $1 - p$. If the local test outputs $-1$, then with probability at least $1 - p$, $D$ does not contain a $\tau$-exceeding point.
\label{lemma:local_test}
\end{lemma}
\begin{lemma}
Let the interval in which $f(x^*)$ lies, as maintained by the algorithm at the beginning of epoch $k$, be denoted by $[a_k, b_k]$. Then $|b_k - a_k| \leq (1 + 2c(\rho_k/d - 1))2^{-\alpha (\rho_k/d-1)}$.
\label{lemma:interval_length}
\end{lemma}

\subsection{Proof of Theorem~\ref{thm:regret}}

% We present the proof of the theorem using the statements of the above lemmas followed by the proofs of the lemmas. \\

For the regret analysis of GP-ThreDS, we write the overall regret as a sum of two terms, $R_1$ and $R_2$.
% each of which evaluates the regret incurred in different time horizons. 
$R_1$ is the regret incurred by the algorithm until the end of the epoch $k_0$, and $R_2$ is the regret incurred by the algorithm after $k_0$ epochs are completed, where $k_0 = \max\{ k : \rho_k \leq \frac{d}{2\alpha} \log T \}$. All the following regret calculations are conditioned on the event that throughout the time horizon, all the random walk modules identify all the target nodes always correctly. We later show that this event occurs with a high probability. \\

We begin with the analysis of $R_1$. To obtain an upper bound on $R_1$, we first obtain the regret incurred at each node and sum that over the different nodes visited by the algorithm in the first $k_0$ epochs.
Since the sampling of the algorithm is independent across different nodes, we can bound the regret incurred at any node $D$ visited by the algorithm during an epoch $k \leq k_0$ independent of others. We denote the discretized version of the domain by $D_g$ and $\xd$ and $\xdg$ are defined as follows: $\xd = \argmax_{x \in D} f(x)$ and $\xdg = \argmax_{x \in D_g} f(x)$.
Recall that the cap on the number of samples in epoch $k$ is defined as
\begin{align*}
	\bar{S}^{(k)}(p) = \min \left\{t \in \N : 2\left( B  + R\sqrt{2(\gamma_{t-1} + 1 + \log(1/p))} \right)(1 + 2\lambda)\sqrt{\frac{|D_g|}{t}} \leq L \Delta_k^{\alpha} \right\} + 1. 
\end{align*}

We focus our attention on any arbitrary node visited during the $k^{\text{th}}$ epoch. Let $N$ denote the random number of queries issued at that node and $\bar{R}(N)$ denote the regret incurred at that node. By the definition of regret, we have,
\begin{align*}
	\bar{R}(N) & =  \sum_{n = 1}^N f(x^*) - f(x_n) \\
	& =  \sum_{n = 1}^N f(x^*) - \tau_k + L \Delta_k^{\alpha} + \tau_k - f(\xdg) -  L \Delta_k^{\alpha} + f(\xdg) - f(x_n)  \\
	& = \underbrace{\left[ \sum_{n = 1}^N f(x^*) - \tau_k +  L \Delta_k^{\alpha}\right]}_{R^{(1)}(N)}  +  \underbrace{\left[ \sum_{n = 1}^N \tau_k -  L \Delta_k^{\alpha} - f(\xdg)\right]}_{R^{(2)}(N)}  +  \underbrace{\left[ \sum_{n = 1}^N f(\xdg) - f(x_n)\right]}_{R^{(3)}(N)}.
\end{align*}
where $x_n$ is the point sampled by the algorithm at the $n^{\text{th}}$ time instant spent at the node. We will bound each of the three terms separately as outlined in Sec.~\ref{sub:regret_analysis}. \\

We begin with bounding the third term, $R^{(3)}(N)$. Notice that it can be bounded in the same way as the regret for IGP-UCB since the sampling is always carried out on the grid by maximizing the UCB score over it.
Since $x_n = \argmax_{x \in D_g} \mu_{n-1}(x) + \beta_n(\delta_0/4T) \sigma_{n-1}(x)$, therefore, with probability at least $1 - \delta_0/4T$, we have
\begin{align*}
    f(\xdg) - f(x_n) & \leq \mu_{n-1}(\xdg) + \beta_n(\delta_0/4T) \sigma_{n-1}(\xdg) - \left(\mu_n(x_n) - \beta_n(\delta_0/4T) \sigma_{n-1}(x_n)\right) \\
    & \leq \mu_{n-1}(x_n) + \beta_n(\delta_0/4T) \sigma_{n-1}(x_n) - \mu_n(x_n) + \beta_n(\delta_0/4T) \sigma_{n-1}(x_n) \\
     & \leq  2\beta_n(\delta_0/4T) \sigma_{n-1}(x_n).
\end{align*}
From Lemma~\ref{lemma:info_gain}, we can conclude that $\sum_{n = 1}^N \sigma_{n-1}(x_n) \leq (1 + 2 \lambda) \sqrt{|D_g| N}$. Using this result along with the bound on $f(\xdg) - f(x_n)$, we obtain
\begin{align*}
    R^{(3)}(N) & =  \sum_{n = 1}^N f(\xdg) - f(x_n) \\
    & \leq \sum_{n = 1}^N 2\beta_n(\delta_0/4T) \sigma_{n-1}(x_n) \\
    & \leq  2\beta_N(\delta_0/4T) \sum_{n = 1}^N \sigma_{n-1}(x_n) \\
    %  & \leq 2\beta_N(\delta_0/4T) \sqrt{|D_g| N } \\
     & \leq 2 \left( B  + R\sqrt{2(\gamma_{N-1} + 1 +  \log(4T/\delta_0))} \right)(1 + 2 \lambda)\sqrt{|D_g| N}.
\end{align*}

To bound the first term, $R^{(1)}(N)$, we relate the maximum number of samples taken at the node to $f(x^*) - \tau_k + L \Delta_k^{\alpha}$ using Lemmas~\ref{lemma:termination_condition} and~\ref{lemma:interval_length}.
Recall the definition of $\bar{S}^{(k)}(p)$. It is defined as 
\begin{align*}
	\bar{S}^{(k)}(p) = \min \left\{t \in \N : 2\left( B  + R\sqrt{2(\gamma_{t-1} + 1 + \log(1/p))} \right)(1 + 2\lambda)\sqrt{\frac{|D_g|}{t}} \leq L \Delta_k^{\alpha} \right\} + 1. 
\end{align*}

This implies that, 
\begin{align*}
    L\Delta_{k}^{\alpha} & \leq 2\left( B  + R\sqrt{2(\gamma_{\bar{S}^{(k)} - 3} + 1 + \log(1/p))} \right)(1 + 2 \lambda)\sqrt{\frac{|D_g|}{\bar{S}^{(k)} - 2}} \\
    \implies 2^{-\alpha\rho_k/d} & \leq \frac{2}{c}\left( B  + R\sqrt{2(\gamma_{\bar{S}^{(k)}} + 1 + \log(1/p))} \right) (1 + 2 \lambda) \sqrt{\frac{3|D_g|}{\bar{S}^{(k)}}} 
\end{align*}

Notice that $f(x^*)$ lies in $[a_k, b_k]$ (under the high probability event on which the analysis is conditioned). Since $\tau_k = (a_k + b_k)/2$, therefore $|f(x^*) - \tau_k| \leq |b_k - a_k|/2$. Using this, we can write $R^{(1)}(N)$ as,

\begin{align*}
    R^{(1)}(N) & = \sum_{n = 1}^N f(x^*) - \tau_k + L \Delta_k^{\alpha}  \\
    & \leq \left(|f(x^*) - \tau_k| + L \Delta_k^{\alpha}\right) N  \\
    & \leq \left(\frac{|b_k - a_k|}{2} + c2^{-\alpha\rho_k/d} \right) N \\
    & \leq \left((1 + 2c(\rho_k/d-1))2^{-\alpha (\rho_k/d-1) - 1} + c2^{-\alpha\rho_k/d} \right) N \\
    & \leq \left((1 + 2c(\rho_k/d-1))2^{-\alpha \rho_k/d} + c2^{-\alpha\rho_k/d} \right) N \\
    & \leq \frac{2N}{c} (1 + c + 2c (\rho_k/d-1)) \left( B  + R\sqrt{2(\gamma_{\bar{S}^{(k)}} + 1 + \log(1/p))} \right) (1 + 2 \lambda) \sqrt{\frac{3|D_g|}{\bar{S}^{(k)}}} \\
    & \leq \frac{2N}{c} (1 + c + \frac{c}{\alpha} \log_2 T ) \left( B  + R\sqrt{2(\gamma_{N} + 1 + \log(1/p))} \right) (1 + 2 \lambda) \sqrt{\frac{3|D_g|}{N}} \\
    & \leq \frac{2}{c} \left(2 + \frac{c}{\alpha} \log_2 T \right) \left( B  + R\sqrt{2(\gamma_{N} + 1 + \log(1/p))} \right)(1 + 2 \lambda) \sqrt{3 |D_g| N},
\end{align*}

where we use Lemma~\ref{lemma:interval_length} in line $4$, definition of $k_0$ in line $7$ and the fact that $N \leq \bar{S}$. Lastly, we consider the second term, $R^{(2)}(N)$. Note that it is trivially upper bounded by zero if $f(\xdg) > \tau_k - L \Delta_k^{\alpha}$. For the case when $f(\xdg) < \tau_k - L \Delta_k^{\alpha}$, we analyze it like $R^{(1)}(N)$ with a different time instant instead of $\bar{S}^{(k)}(p)$. Define $t_1$ as
\begin{align*}
	t_1 = \min \left\{t \in \N : 2\left( B  + R\sqrt{2(\gamma_{t-1} + 1 +  \log(4T/\delta_0))} \right)(1 + 2 \lambda)\sqrt{\frac{|D_g|}{t}} \leq \tau_k - L \Delta_k^{\alpha} - f(\xdg) \right\}.
\end{align*}
From Lemma~\ref{lemma:termination_condition}, we know that $\Pr(N > t_1) \leq \dfrac{\delta_0}{4T}$. Therefore, with probability at least $1 - \dfrac{\delta_0}{4T}$, we have $N \leq t_1$. Conditioning on this event and using a similar sequence of arguments as used in proof of $R^{(1)}(N)$, we can write

\begin{align*}
    \tau_k - L \Delta_k^{\alpha} - f(\xdg) \leq 2\left( B  + R\sqrt{2(\gamma_{t_1} + 1 +  \log(4T/\delta_0))} \right)(1 + 2 \sigma)\sqrt{\frac{2 |D_g|}{t_1}}.
\end{align*}

Thus with probability at least $1 - \dfrac{\delta_0}{4T}$, we have,
\begin{align*}
    R^{(2)}(N) & =\sum_{n = 1}^N \tau_k - L \Delta_k^{\alpha} - f(\xdg) \\
    & \leq \left( \tau_k - L \Delta_k^{\alpha} - f(\xdg) \right) N \\
    & \leq 2N\left( B  + R\sqrt{2(\gamma_{t_1} + 1 +  \log(4T/\delta_0))} \right)(1 + 2 \lambda)\sqrt{\frac{2 |D_g|}{t_1}} \\
    & \leq 2 \left( B  + R\sqrt{2(\gamma_N + 1 +  \log(4T/\delta_0))} \right)(1 + 2 \lambda)\sqrt{2|D_g| N}. 
\end{align*}

% \begin{align*}
%     R^{(1)}(N) & =\sum_{n = 1}^N \tau_k - L \Delta_k^{\alpha} - f(\xdg) \\
%     & \leq \left( \tau_k - L \Delta_k^{\alpha} - f(\xdg) \right) N \\
%     & \leq 2N\left( B  + R\sqrt{2(\gamma_{t_1} + 1 +  \log(4T/\delta_0))} \right)\sqrt{\frac{16 \gamma_{t_1}}{t_1}} \\
%     & \leq 2 \left( B  + R\sqrt{2(\gamma_N + 1 +  \log(4T/\delta_0))} \right)\sqrt{16 \gamma_{N} N}. 
% \end{align*}

On combining all the terms, we can conclude that $\bar{R}(N)$ is $O( \log T \sqrt{N (\gamma_N + \log(T/\delta_0)})$.
% On combining all the terms, we can conclude that $\bar{R}(N)$ is $O(\gamma_N \log T \sqrt{N})$.
To compute $R_1$, we just need to evaluate the total number of nodes visited by the algorithm in the first $k_0$ epochs. Using Lemma~\ref{lemma:number_of_nodes}, we can conclude that with probability at least $1 - {\delta_0}/{4T}$, one iteration of random walk would have visited less than $\displaystyle \frac{1}{2(p - 1/2)^2} \log \left( \frac{4dT}{\delta_0}\right)$ nodes. Therefore, throughout the algorithm, all iterations of random walks would have visited less than $\displaystyle \frac{1}{2(p - 1/2)^2} \log \left( \frac{4dT}{\delta_0}\right)$ nodes  with probability at least $1 - \delta_0/4$.\\

Let $L_m$ denote the number of nodes at depth $md$ of tree $\cT_0$ for $m = 1, 2, \dots, k_0$ that contain a point $x$ such that $f(x) \geq \tau_m - c2^{-\alpha \rho_m/d + 1}$. Therefore $L_m$ denotes an upper bound on the number of target nodes for epoch $m$. Let $L_0 = \max_{1 \leq i \leq k_0} L_i$. Using the upper bound on the number of nodes visited during on iteration of RWT,  we can conclude that the algorithm would have visited less than $\displaystyle K = \frac{k_0 L_0}{(p - 1/2)^2} \log \left( \frac{4dT}{\delta_0}\right)$ nodes in the first $k_0$ epochs with probability at least $1 - \delta_0/4$. To bound $k_0$, note that the update scheme of the interval $[a_k, b_k]$ (and consequently $\tau_k$) ensures that the algorithm does not spend more than $2$ epochs at any specific depth of the tree. This implies that $k \leq 2 \rho_k/d$. Thus, $k_0 \leq \frac{1}{\alpha} \log_2 T$. \\

Let $N_j$ denote the random number of queries at node $j$ visited during the algorithm and $\bar{R}_j(N_j)$ denote the associated regret for $j = 1,2, \dots, K$. Therefore, for some constant $C_0$, independent of $T$, we have, 
\begin{align*}
	R_1 & \leq \sum_{j = 1}^K \bar{R}_j(N_j) \\
	& \leq C_0 \log T \sum_{j = 1}^K  \sqrt{N_j (\gamma_{N_j} + \log(T/\delta_0) )} \\
	& \leq C_0 \log T \sqrt{\gamma_{T} + \log(T/\delta_0)}  \sum_{j = 1}^K \sqrt{N_j} \\
	& \leq C_0  \log T \sqrt{\gamma_{T} + \log(T/\delta_0)} \cdot \sqrt{K \sum_{j = 1}^K N_j} \\
	& \leq C_0  \log T \sqrt{\gamma_{T} + \log(T/\delta_0)} \cdot \sqrt{KT} \\
	& \leq C_0 \sqrt{\frac{T \log(T) L_0}{2(p - 1/2)^2} \log \left( \frac{4dT}{\delta_0} \right)} \cdot \sqrt{\gamma_{T} + \log(T/\delta_0)} \cdot \log T.
\end{align*}

Therefore, $R_1$ is $O(\sqrt{T \gamma_T} \log T \sqrt{\log T \cdot \log(T/\delta_0)} )$. \\

We now focus on bounding $R_2$. Let $D$ represent a node being visited after $k_0$ epochs and $\xd = \argmax_{x \in D} f(x)$. The instantaneous regret at time instant $t$ can be written as 
\begin{align*}
    r_t & = f(x^*) - f(x_t) \\
    & = [f(x^*) - f(\xd)] + [f(\xd) - f(x_t)]. 
\end{align*}
We bound both the expressions, $f(x^*) - f(\xd)$ and $f(\xd) - f(x_t)$ separately for any such node. We begin with the second expression.
% beginning with the latter.
After $k_0$ epochs, all the high-performing nodes being considered by the algorithm would be at a depth of at least $\frac{d}{2 \alpha}\log_2 T$ in the original infinite binary tree. This implies that the length of the edges of the cuboid corresponding to the nodes would be smaller than $ T^{-1/(2 \alpha)}$. Consequently, no two points in any such node would be more than $\sqrt{d}T^{-1/(2 \alpha)}$ apart. Therefore, $\displaystyle f(\xd) - f(x_t) \leq L \sqrt{d^{\alpha}/T}$, where $x_t$ is a point sampled at time instant $t$ after $k_0$ epochs have been completed. To bound the first expression, notice that $f(\xd) \in [a_{k_0 + 1}, b_{k_0+1}]$ for all nodes visited after $k_0$ epochs have been completed. This follows from the construction of intervals $[a_k, b_k]$. Since $f(x^*)$ also lies in $[a_{k_0 + 1}, b_{k_0+1}]$ (under the high probability event), we have, $f(x^*) - f(\xd) \leq |b_{k_0+1} - a_{k_0+1}| \leq (1 + 2c(\rho_{k_0+1}/d-1))2^{-\alpha (\rho_{k_0+ 1}/d -1)}$ for any node visited after $k_0$ epochs. Therefore, we can bound the instantaneous regret as
\begin{align*}
    r_t & = [f(x^*) - f(\xd)] + [f(\xd) - f(x_t)] \\
    & \leq (1 + 2c(\rho_{k_0+1}/d-1))2^{-\alpha (\rho_{k_0+ 1}/d -1)} + L \sqrt{\frac{d^{\alpha}}{T}} \\
    & \leq 2(2 + \frac{c}{2\alpha} \log_2 T)\sqrt{\frac{1}{T}} + L \sqrt{\frac{d^{\alpha}}{T}}.
\end{align*}
If $T_{R_2}$ denotes the samples taken by the algorithm after completing $k_0$ epochs, then $R_2$ can be bounded as 
\begin{align*}
    R_2 \leq \frac{T_{R_2}}{\sqrt{T}} \left(2 \left(2 + \frac{c}{2\alpha} \log_2 T\right) + L \sqrt{d^{\alpha}} \right).
\end{align*}
Noting that $T_{R_2} \leq T$, we have that $R_2$ is $O(\sqrt{T} \log T)$. On adding the bounds on $R_1$ and $R_2$, we obtain that the regret incurred by the algorithm is $O(\sqrt{T \gamma_T} \log T \sqrt{\log T \cdot \log(T/\delta_0)} )$, as required. \\

We now show that this bound holds with high probability. Firstly, we had obtained a bound on $R^{(3)}(N)$ for a node $D \subseteq \cX$ by conditioning on the event that $|f(x) - \mu_{t-1}(x)| \leq \beta_{t}(\delta_0/4T)\sigma_{t-1}(x)$ holds for all $x \in D$ and $t \geq 1$. Since the probability that event occurs is at least $1 - \delta_0/4T$, the bound on $R^{(3)}(N)$ holds simultaneously for all nodes visited throughout the time horizon with a probability of at least $1 - \delta_0/4$.
Similarly, to obtain a bound on $R^{(2)}(N)$ for any node $D \subseteq \cX$, we had conditioned the analysis on another event which holds with a probability of at least $1 - \delta_0/4T$. Therefore, the bound on $R^{(2)}(N)$ holds simultaneously for all nodes visited by the algorithm with a probability of at least $1 - \delta_0/4$. We also note that while using Lemma~\ref{lemma:number_of_nodes} to bound the number of nodes visited by the algorithm, we had conditioned the analysis on another event (that bounded the number of nodes visited in an iteration of RWT) that holds with a probability of at least $1 - \delta_0/4$ (See Sec.~\ref{sub:proof_of_lemma_number_of_nodes}). Lastly, since we assume that the algorithm always identifies all the target nodes correctly, we also need to account for the probability that this is true. From the error analysis of RWT as described in Section~\ref{sub:proof_of_lemma_number_of_nodes}, we note that every target node is identified correctly with a probability of at least $1 - \delta_0/4T$. Therefore, using a probability union bound, the algorithm identifies all the target nodes correctly with a probability of no less than $1 - \delta_0/4$. Combining all the above observations, we can conclude that the above obtained regret bound holds with a probability of at least $1 - \delta_0$, as required.

\subsection{Proof of Lemma~\ref{lemma:info_gain}}
\label{subsub:proof_sum_sigmas}

We consider a domain $D \subseteq \cX$ and its discretization $D_g$ that contains $|D_g|$ number of points. Let the points be indexed from $1$ to $|D_g|$ and let $n_i$ denote the number of times the $i^{\text{th}}$ point was chosen in the set of sampled points $\{x_1, x_2, \dots, x_t\}$. Let $\cI = \{i : n_i > 0\}$ and $|\cI|$ denote the number of elements in $\cI$. Consider the $i^{\text{th}}$ point, denoted by $x^{(i)}$, and let $1 \leq t_1 <  t_2 <  \dots < t_{n_i} \leq t$ denote the time instances when the $i^{\text{th}}$ point is sampled, that is, at time $t_j$, it is sampled for the $j^{\text{th}}$ time, for $j = 1,2, \dots, n_i$. Clearly, we have $\sigma_{t_1 -1}(x_{t_1}) = \sigma_{t_1 - 1}(x^{(i)}) \leq k(x^{(i)}, x^{(i)}) \leq 1$. For all $2 \leq j \leq n_i$, at time instant $t_j$, $x^{(i)}$ has been sampled for $j - 1$ times before $t_j$.
Using Proposition 3 from~\cite{Shekhar2018}, we have $\sigma_{t_j - 1}(x_{t_j}) = \sigma_{t_j -1}(x^{(i)}) \leq \dfrac{\lambda}{\sqrt{j -1}}$. This can be interpreted as bounding the standard deviation by only the contribution coming from the noisy observations. We would like to emphasize that we are using the Proposition 3 for the \emph{surrogate} GP-model adopted for the optimization. While the actual noise is indeed $R$-sub-Gaussian, we are applying the Proposition 3 bearing in mind the \emph{fictitious} Gaussian noise assumption for our \emph{surrogate} model. \\

Thus for each point in $\cI$, the contribution to the sum is upper bounded by $\displaystyle 1 + \lambda \sum_{j = 1}^{n_i- 1} j^{-1/2}$. Thus, we have, 
\begin{align*}
    \sum_{s = 1}^t \sigma_{s-1}(x_s) & \leq \sum_{i \in \cI} \left( 1 +  \lambda \sum_{j = 1}^{n_i-1} \frac{1}{\sqrt{j}}  \right) \\
    & \leq \sum_{i \in \cI}\left( 1 +   \lambda \int_{0}^{n_i-1} \frac{1}{\sqrt{z}} \ dz \right)\\
    & \leq  \sum_{i \in \cI}\left( 1 +  2\lambda \sqrt{n_i-1} \right) \\
    & \leq (1 + 2\lambda) \sum_{i \in \cI} \sqrt{n_i}  \\
    & \leq (1 + 2\lambda)  |\cI|  \sqrt{ \frac{1}{|\cI|} \sum_{i \in \cI} n_i} \\
    & \leq (1 + 2\lambda) \sqrt{ |\cI| t}.
\end{align*}
In the fifth step, we have used Jensen's Inequality. Noting that $|\cI| \leq |D_g|$, we obtain the required result.

\subsection{Proof of Lemma~\ref{lemma:termination_condition}} % (fold)
\label{sub:proof_of_lemma_termination_condition}

Consider the performance of the local test on a domain $D \subseteq \cX$ with a threshold $\tau$. The discretized version of the domain is denoted by $D_g$. As before, we use the following notation throughout the proof of this lemma,. Let $\xd = \argmax_{x \in D} f(x)$, $\xdg = \argmax_{x \in D_g} f(x)$, $\hat{x}_t = \argmax_{x \in D_g} \mu_{t-1}(x) + \beta_t(p)\sigma_{t-1}(x) $ and let $\bar{x}_t = \argmax_{x \in D_g} \mu_{t-1}(x) - \beta_{t}(p)\sigma_{t-1}(x) $. Lastly, recall that the termination time is defined as
\begin{align*}
	\bar{S}(\delta_2, \Delta_f) = \min \left\{t \in \N : 2\left( B  + R\sqrt{2(\gamma_{t-1} + 1 + \log(1/\delta_2))} \right)(1 + 2\lambda) \sqrt{\frac{|D_g|}{t}} \leq \Delta_f \right\} + 1. 
\end{align*}

Let us consider the case when $f(\xdg) < \tau - L \Delta^{\alpha} - \Delta_f$ and let $N$ denote the random number of samples taken in a sequential test without a cap on the total number of samples. We first make the following observation about the posterior variance at the point to be sampled at $t$, $x_t$, and $\hat{x}_t$. From the definitions of $x_t$ and $\hat{x}_t$, we have, % We have the following relations.
\begin{align*}
	\mu_{t - 1}(x_t) + \beta_t(\delta_0/4T)\sigma_{t-1}(x_t) & \geq \mu_{t - 1}(\hat{x}_t) + \beta_t(\delta_0/4T)\sigma_{t-1}(\hat{x}_t) \\
	\mu_{t - 1}(\hat{x}_t) + \beta_t(p)\sigma_{t-1}(\hat{x}_t) & \geq \mu_{t - 1}(x_t) + \beta_t(p)\sigma_{t-1}(x_t)
\end{align*}
On adding the two, we obtain that $\sigma_{t-1}(\hat{x}_t) \leq \sigma_{t-1}(x_t)$. Note that this holds for all $t$. Next, we define the event $E$ as $|f(x) - \mu_{t-1}(x)| \leq  \beta_t(\delta_2) \sigma_{t-1}(x)$ being true for all $x \in D$ and $t \geq 1$. From~\cite[Theorem 2]{Chowdhury2017}, we know that the probability of $E$ is at least $1 - \delta_2$. Let $E^c$ denote the complement of the event $E$. Using the event $E$, we evaluate the probability that the local test queries more than $n$ points. The probability that $N > n$ can be written as follows,
\begin{align*}
    \Pr(N > n) & \leq \Pr \left( \left\{ \forall t \leq n : \mu_{t-1}(\hat{x}_t) + \beta_t(p)\sigma_{t-1}(\hat{x}_t)  \geq \tau - L \Delta^{\alpha}  \right\} \right) \\
    & \leq \Pr \left( \left\{ \forall t \leq n : \mu_{t-1}(\hat{x}_t) + \beta_t(p)\sigma_{t-1}(\hat{x}_t)  \geq \tau - L \Delta^{\alpha}  \right\} | E \right) \Pr(E) + \\
    & \ \ \ \ \ \Pr \left( \left\{ \forall t \leq n : \mu_{t-1}(\hat{x}_t) + \beta_t(p)\sigma_{t-1}(\hat{x}_t)  \geq \tau - L \Delta^{\alpha}  \right\} | E^c \right) \Pr(E^c) \\ 
    & \leq \Pr \left( \sum_{t = 1}^n \mu_{t-1}(\hat{x}_t) + \beta_t(p)\sigma_{t-1}(\hat{x}_t)  \geq \sum_{t = 1}^n (\tau - L \Delta^{\alpha})  \bigg| E \right)  +  \Pr(E^c) \\ 
    & \leq \Pr \left( \sum_{t = 1}^n f(\hat{x}_t) + \beta_t(\delta_2)\sigma_{t-1}(\hat{x}_t) + \beta_t(p)\sigma_{t-1}(\hat{x}_t)  \geq \sum_{t = 1}^n (\tau - L \Delta^{\alpha})  \bigg| E \right)  +  \delta_2 \\ 
    & \leq \Pr \left( \sum_{t = 1}^n f(\xdg) + 2\beta_t(\delta_2)\sigma_{t-1}(\hat{x}_t) \geq \sum_{t = 1}^n (\tau - L \Delta^{\alpha})   \bigg| E \right)  +  \delta_2 \\ 
     & \leq \Pr \left( \sum_{t = 1}^n 2\beta_t(\delta_2)\sigma_{t-1}(x_t) \geq \sum_{t = 1}^n (\tau - f(\xdg)- L \Delta^{\alpha})   \bigg| E \right)  +  \delta_2 \\ 
\end{align*}

To bound the first term on the RHS, we make use Lemma~\ref{lemma:info_gain}.

Therefore, we have 
\begin{align*}
	\frac{1}{n} \sum_{t = 1}^n 2\beta_t(\delta_2)\sigma_{t-1}(x_t) & \leq \frac{2\beta_n(\delta_2)}{n} \sum_{t = 1}^n \sigma_{t-1}(x_t) \\
	& \leq \frac{2\beta_n(\delta_2)}{n} (1 + 2\lambda)\sqrt{|D_g| n}  \\
	& \leq 2\beta_n(\delta_2) (1 + 2\lambda) \sqrt{\frac{|D_g|}{n}} \\
	& \leq \Delta_f < \tau  - f(\xdg) - L \Delta^{\alpha}.
\end{align*}

This implies that the first term on RHS goes to zero for $n \geq \bar{S} - 1$ implying that the probability that the local test takes more than $\bar{S}$ samples when $f(\xdg) < \tau - L \Delta^{\alpha} - \Delta_f$ is less than $\delta_2$. This implies if the local test has reached the termination condition then with probability atleast $1 - \delta_2$, we have that $f(\xdg) > \tau - L \Delta^{\alpha} - \Delta_f$. We can carry out a similar analysis for the case when $f(\xdg) > \tau + \Delta_f$ to obtain the statement of the lemma.

% subsection proof_of_lemma_termination_condition (end)

\subsection{Proof of Lemma~\ref{lemma:number_of_nodes}} % (fold)
\label{sub:proof_of_lemma_number_of_nodes}

The proof of this lemma is mainly based on the analysis of random walk on a binary tree. This analysis is similar to the one described in~\cite{Wang2018}. We reproduce a slightly different version of the proof that is more focused on finding a high probability bound on the number of nodes visited in the random walk. In this proof, we consider a binary tree of depth $d$, denoted by $\hat{\cT}$, to represent the tree considered in the random walk. We index the leaf nodes from $1$ to $n$ where $n = 2^d$. Throughout this proof, we refer to the high-performing nodes as target nodes. We begin with the case of a single target and then extend the proof for the case of multiple targets. \\

WLOG, we consider the single target node to be the leaf node indexed as $1$. We divide the tree $\hat{\cT}$ into a sequence of sub-trees denoted by $\hat{\cT}_0, \hat{\cT}_1, \dots$ for $i =0, 1, 2, \dots d$ which are defined as follows. Consider the nodes on the path joining the root node to the target node. Such a path is unique as the underlying graph is a tree. Let $v_i$ denote the node on this path that is at a distance of $i$ from the target node. The distance between two nodes is defined as the length of the path connecting those two nodes. $\hat{\cT}_i$ is defined to be tree that contains the node $v_i$ along with the sub-tree rooted at the child that does not contain the target node. \\ 

This construction is similar to the one outlined in~\cite{Wang2018}. Also, $\hat{\cT}_{0}$ corresponds to the target node. Since the random walk is biased towards the minimizer, given the construction of $\hat{\cT}_i$, the probability that random walk is still in one of such sub-trees would decrease with time. To formalize this idea, we consider the last passage times of any sub-tree $\hat{\cT}_{i}$ for $1\leq i \leq d$. Let $\tau_i$ denote the last passage time to $\hat{\cT}_i$. \\

We begin with the analysis for $\tau_d$. This problem of random walk on $\hat{\cT}_{d}$ can be mapped to the problem of a random walk on the set $S = \{ -1, 0, 1, 2, \dots d \}$. If each non-negative integer is mapped to the subset of nodes at the corresponding depth in the sub-tree, then our random walk on $\cT_{d}$ between different levels is equivalent to the random walk on these integers. Note that since the target node is not contained in this sub-tree, all nodes at the same depth are identical in terms of distance to the target node. In particular, they all are equally far away from exiting the tree and therefore can be abstracted into single node. This abstraction is precisely what leads to the equivalence between the two problems. Under this setup, the root node is mapped to $0$ and the sub-tree containing the target node is mapped to $-1$, indicating an exit from the sub-tree $\hat{\cT}_d$. \\

We begin the random walk at integer $0$ where escaping the tree is equivalent to arriving on the integer $-1$. For the random walk to arrive on $-1$, it would have to take greater number of steps in the negative direction than it took in the positive one. Also, since the probability of moving along the negative direction is at least $1-p$, we can write,
\begin{align*}
    \mathbb{P}(\tau_{d} > n) \leq \mathbb{P}(Z \leq n/2),
\end{align*}
where $Z \sim \mathrm{Bin}(n, p)$ is a Binomial random variable. Therefore, we have
\begin{align*}
    \mathbb{P}(\tau_{d} > n) \leq \exp(-2(p -1/2)^2 n).
\end{align*}
On account of the underlying symmetry, we can conclude that this bound holds for all $i$. Therefore, we have $\mathbb{P}(\tau_{i} > n) \leq \exp(-2(p -1/2)^2 n)$ for all $i = 0, 1, \dots, d$. \\

For the case of multiple target nodes, we can construct a similar set of sub-graphs and conclude the same result for those sub-graphs. Note that we redefine these set for every different iteration of the random walk when it restarts after detecting a target node. Consider the case when there are $L$ target nodes. We begin with considering the first iteration of the random walk. For each target node $j = 1, 2, \dots, L$, we define a sequence of sub-trees $\cT_{i}^{(j)}$ for $i = \{0,1, \dots, d \}$ exactly in the same manner as we did in the previous case. That is, $\cT_{i}^{(j)}$ would be a tree consisting of the node that lies on the path between the target node $j$ and the root node and is at a distance of $i$ from the target node, along with child that does not contain the target node $j$. By definition, the sub-trees $\cT_{i}^{(j)}$ are not disjoint for different values of $j$. Using these sub-trees, we define a partition of the binary tree denoted by the sub-graphs $\hat{\cT}_{i}'$ for $i = \{0,1, \dots, d \}$ as follows. If $\cV$ denotes the set of all nodes on the binary tree, then for each $v \in \cV$, we define $v(j) = \{i : v \in \cT_{i}^{(j)}\}$. Therefore, $v(j)$ denotes the index of the sub-tree corresponding to the target node $j$ to which the node $v$ belongs. From the construction of $\cT_{i}^{(j)}$, it follows that $v(j)$ is unique for each $v \in \cV$. Using this, we define 
\begin{align*}
	\hat{\cT}_{i}' = \{ v \in \cV :  \min_{j} v(j) = i \}
\end{align*}
In other words, $\hat{\cT}_{i}'$ consists of all the nodes such that there is at least one target node $j$ for which it belongs to $\cT_{i}^{(j)}$.  \\

The motivation is that if the random walk escapes $\hat{\cT}_i'$ in the correct direction then it has moved closer to at least one of the target nodes. It is not difficult to note that this is exactly how the sub-trees $\hat{\cT}_i$ were designed in the previous proof. The only difference between the two cases is that $\hat{\cT}_{i}'$ is designed to accommodate the presence of multiple target nodes where all the target nodes have the same level of preference for the random walk. In a similar vein to the case of a single target, we define $\tau_i'$ as the last passage time to $\hat{\cT}_{i}'$ for $i = \{0,1, \dots, d\}$. \\

Leveraging the similarity of definitions of $\hat{\cT}_{i}'$ and $\hat{\cT}_{i}$ along with the agnosticism of the random walk to the target node, we can use exactly the same analysis as for the single target case to conclude that 
\begin{align*}
    \mathbb{P}(\tau_{i}' > n) \leq \exp(-2(p -1/2)^2 n).
\end{align*}

We let $M$ denote the random number of steps taken by one iteration of random walk before termination. Therefore, we can write,
\begin{align*}
	\Pr(M > r) & \leq \Pr \left( \bigcup_{i = 0}^d \{ \tau_{i}' > r \}\right) \\
	& \leq \sum_{i = 0}^d \Pr \left(  \tau_{i}' > r \right) \\
	& \leq \sum_{i = 0}^d  \exp(-2(p -1/2)^2 r) \\
	& \leq d\exp(-2(p -1/2)^2 r) 
\end{align*}

Using the above relation, we can conclude that one iteration of the random walk will take less than $\dfrac{1}{2(p - 1/2)^2} \log \left( \dfrac{d}{\delta_1}\right)$ with probability at least $1 - \delta_1$, as required.\\

This also helps bound the probability of error in the random walk. If $M_1$ denotes the number of non-target leaf nodes visited in the random walk, then the probability that a target node is identified incorrectly is less than $M_1 \delta_2$, where $\delta_2$ is the error probability for the leaf test. Using the above bound on $M_1$ with $\delta_1 = \delta_0/4T$ along with the value of $\delta_2$ as specified by the algorithm (See Appendix~\ref{sub:RWT}), we conclude that an iteration of random walk identifies a target node correctly with probability at least $1 - \delta_0/4T$.

\subsection{Proof of Lemma~\ref{lemma:constant_nodes}}
\label{sub:proof_of_lemma_constant_nodes}

A key idea in the proof of the lemma is to establish that for the choice of parameters used in GP-ThreDS, the rate at which domain shrinks matches the rate at which the discretization gets finer. Let $D$ be a node visited by the algorithm during epoch $k$ and $D_g$ be its associated discretization such that 
\begin{align*}
    \sup_{x \in D} \inf_{y \in D_g} \| x - y \| \leq \Delta_k.
\end{align*}
More specifically, $D$ refers to the subset of the domain corresponding to the node visited by the algorithm.

Note that using the definition of a covering set, we can conclude that $D_g$ is a  $\Delta_k$-cover of $D$. Then, using the bounds on the covering number of a hypercube in $\R^d$ \cite{Shalev2014}, we have that $|D_g|$ is $O(\mathrm{vol}(D)\Delta_k^{-d})$. Since $D$ is a node visited during epoch $k$ of the algorithm, it lies at a depth of at least $\rho_k - d$ on the infinite depth binary tree constructed on the domain. From the construction of the binary tree, we note that the lengths of nodes in all the dimensions get halved every $d$ steps.
Thus, the lengths of the edges of the cuboid corresponding to $D$ are less than $2^{-\rho_k/d + 1}$. Consequently, $\mathrm{vol}(D)$ is $O(2^{-\rho_k})$. On substituting this value in the bound for $|D_g|$ along with $\Delta_k = (c/L)^{1/\alpha}2^{-\rho_k/d}$, we obtain $|D_g|$ is $O(1)$, independent of $k$ (and thus $t$). The exponential dependence of $|D_g|$ on $d$ also immediately follows from the above analysis. \\

As mentioned in Sec.~\ref{sub:computational_complexity}, the proof of Theorem~\ref{thm:computational_complexity} follows from this lemma. Since only a constant number of UCB scores have to be evaluated at every time instant, the matrix inversion becomes the dominant cost resulting in a worst-case computational cost of $O(t^3)$ at time $t$. Consequently, this results in worst-case overall computational complexity of $O(T^4)$.

\subsection{Proof of Lemma~\ref{lemma:local_test}} % (fold)
\label{sub:proof_of_lemma_local_test}

For the analysis of the local test, we consider several cases based on the maximum value of the function on the grid and consider the results obtained in each one of them. \\

We consider the performance of the local test on a node corresponding to $D \subseteq \cX$ visited by the random walk during epoch $k$. The discretized version of the domain is denoted by $D_g$. Recall that during epoch $k$, the closest point in $D_g$ from any point $x \in D$ is at a distance less than $\Delta_k$. 
% = (c/L)^{1/\alpha}2^{-k}$. 
We define $\xd, \xdg, \hat{x}_t$ and $\bar{x}_t$ in the same way as in the proof of Lemma~\ref{lemma:termination_condition} (Appendix~\ref{sub:proof_of_lemma_termination_condition}).

The cap on the number of samples in epoch $k$ is given as
\begin{align*}
	\bar{S}^{(k)}(p) = \min \left\{t \in \N : 2\left( B  + R\sqrt{2(\gamma_{t-1} + 1 + \log(1/p))} \right)(1 + 2 \lambda)\sqrt{\frac{|D_g|}{t}} \leq L \Delta_k^{\alpha} \right\} + 1. 
\end{align*} 

Similar to the proof of Lemma~\ref{lemma:termination_condition} (Appendix~\ref{sub:proof_of_lemma_termination_condition}), we define the event $E$ as the inequality $|f(x) - \mu_{t-1}(x)| \leq  \beta_t(p) \sigma_{t-1}(x)$ being true for all $x \in D$ and $t \geq 1$. We know this event occurs with a probability of at least $1 - p$. For the following analysis, we assume that event $E$ occurs. Consider the following scenarios based on the value of $f(\xdg)$.
\begin{itemize}
    \item $f(\xdg) > \tau_k + L \Delta_k^{\alpha}$: \\
    From the results obtained in the proof of Lemma~\ref{lemma:termination_condition}, we know that the local test will not terminate. Also notice that the local test cannot return $-1$ as
    \begin{align*}
        \mu_{t-1}(\hat{x}_t) + \beta_t(p)\sigma_{t-1}(\hat{x}_t) & \geq \mu_{t-1}(\xdg) + \beta_t(p)\sigma_{t-1}(\xdg) \\
        & \geq f(\xdg) \\
        & > \tau_k - L\Delta_k^{\alpha}.
    \end{align*}
    Therefore, the local test will always return $+1$.
    \item $\tau_k + L \Delta_k^{\alpha} \geq f(\xdg) \geq \tau_k$: \\
    Similar to the previous case, we can conclude that the local test will never return $-1$. It may return $+1$ or terminate.
    \item $\tau_k  > f(\xdg) > \tau_k - L \Delta_k^{\alpha}$: \\
    Again, similar to the previous cases, the local test will never return $-1$. For this case, we also have,
    \begin{align*}
        \mu_{t-1}(\bar{x}_t) - \beta_t(p)\sigma_{t-1}(\bar{x}_t) & \leq f(\bar{x}_t) \\
        & \leq f(\xdg) \\
        & < \tau_k.
    \end{align*}
    Therefore, the local test will also never return $+1$ (before termination) implying it will always terminate. 
    \item $\tau_k - L \Delta_k^{\alpha}  \geq f(\xdg) \geq \tau_k - 2L \Delta_k^{\alpha}$: \\
    Similarly, the local test will not return $+1$ (before termination). It may return $-1$ or terminate.
    \item $\tau_k - 2L \Delta_k^{\alpha}  > f(\xdg)$: \\
    % Since the local test will not return $+1$ and f
    From the results obtained in Sec.~\ref{sub:proof_of_lemma_termination_condition} we can show the local will neither terminate nor return $+1$, implying that the local test will always return $-1$.
\end{itemize}
From the above analysis, one can directly obtain the statement of the lemma. If $D$ is high-performing with respect to the threshold $\tau_k$, then $f(\xd) > \tau_k$ implying that $f(\xdg) > \tau_k - L \Delta_{k}^{\alpha}$. If $f(\xdg) > \tau_k - L \Delta_{k}^{\alpha}$, then the local test will output $+1$ whenever event $E$ occurs, i.e., with a probability of at least $1- p$. Similarly, if the local test outputs $-1$ when $E$ has occurred, we know that $f(\xdg) < \tau_k - L \Delta_{k}^{\alpha}$, implying $f(\xd) < \tau_k$ and hence $D$ is not high-performing w.r.t. the threshold $\tau_k$, as required. \\

However, we would like to point out that when $\tau_k - L \Delta_k^{\alpha}  > f(\xdg) > \tau_k - 2L \Delta_k^{\alpha}$, the test may output $-1$ or terminate, in which case it returns a $+1$ and accept the current node. This happens because of the conservative nature of the local test. In order to avoid missing nodes that contain a point with a function value greater than $\tau_k$, the local test sometimes accepts nodes nodes like these which have a point with a function value greater than $\tau_k - c2^{-\alpha\rho_k/d + 1}$ but not greater than $\tau_k$. This explains the reason behind the particular choice of values used in the update policy of $\tau_k$.

\subsection{Proof of Lemma~\ref{lemma:interval_length}} % (fold)
\label{sub:proof_of_lemma_interval_length}

We prove the statement of the lemma using induction.
Recall that $[a_k, b_k]$ denotes the interval to which $f(x^*)$ is likely to belong at the beginning of epoch $k$. %Note that 
For the base case, for the LHS we have $|b_1 - a_1| = |b-a| = 1$. Since $\rho_1 = d$, the RHS also evaluates to $1$ verifying the base case. Let us assume that the relation $|b_k - a_k| \leq (1 + 2c(\rho_k-d)/d)2^{-\alpha (\rho_k/d - 1)}$ is true for some $k \geq 1$. 

In the event that the algorithm does not find any $\tau_k$-exceeding point, we set $a_{k+1} = a_k - (b_k - a_k)/2$, $b_{k+1} = b_k - (b_k - a_k)/2$ and $\rho_{k+1} = \rho_k$. Thus, we have, 
\begin{align*}
    |b_{k+1} - a_{k+1}| & = \left| b_k - \frac{b_k - a_k}{2} - a_k +\frac{b_k - a_k}{2} \right| \\
    & = |b_k - a_k| \\
    & \leq (1 + 2c(\rho_k-d)/d)2^{-\alpha (\rho_k/d - 1)} \\
    & \leq (1 + 2c(\rho_{k+1}-d)/d)2^{-\alpha (\rho_{k+1}/d - 1)},
\end{align*}
as required. Now, if the algorithm finds a $\tau_k$-exceeding point, we set $a_{k+1} = \tau_k - c2^{-\alpha \rho_k/d + 1}$, $b_{k+1} = b_k$ and $\rho_{k+1} = \rho_k + d$. Thus, we have, 
\begin{align*}
    |b_{k+1} - a_{k+1}| & = |b_k - \tau_k + c2^{-\alpha \rho_k/d  + 1}| \\
    & \leq \frac{1}{2}|b_k - a_k| + c2^{-\alpha\rho_k/d + 1} \\
    & \leq \frac{1}{2}\left(1 + \frac{2c(\rho_k-d)}{d}\right)2^{-\alpha (\rho_k/d-1)} + c2^{-\alpha\rho_k/d + 1} \\
    & \leq \left(1 + \frac{2c(\rho_k-d)}{d}\right)2^{-\alpha \rho_k/d} + 2c2^{-\alpha\rho_k/d} \\
    & \leq \left(1 + 2c \left(\frac{(\rho_k + d) - d}{d} \right)\right)2^{-\alpha (\rho_{k+1}/d - 1)} \\
    & \leq \left(1 + 2c \left(\frac{\rho_{k+1}- d}{d} \right)\right)2^{-\alpha (\rho_{k+1}/d - 1)} \\
    & \leq \left(1 + \frac{2c(\rho_{k+1} - d)}{d} \right)2^{-\alpha (\rho_{k+1}/d - 1)},
\end{align*}
as required. This completes the proof.

% subsection proof_of_lemma_interval_length (end)

% subsection proof_of_theorem_regret (end)